%% file: preprint.tex
\documentclass{article}

\PassOptionsToPackage{numbers, compress}{natbib}
\usepackage[preprint]{neurips_2026}

\usepackage[utf8]{inputenc}%
\usepackage[T1]{fontenc}%
\usepackage{hyperref}%
\usepackage{url}%
\usepackage{booktabs}%
\usepackage{amsfonts}%
\usepackage{nicefrac}%
\usepackage{microtype}%
\usepackage[table]{xcolor}%
\definecolor{olivehighlight}{RGB}{238,250,248}
\definecolor{oliveintegrated}{RGB}{253,246,249}

\usepackage{graphicx}
\usepackage{float}
\usepackage{wrapfig}
\usepackage{pgfplots}
\pgfplotsset{compat=1.18}
\usepackage{multirow}
\usepackage{enumitem}
\usepackage{amsmath}
\usepackage{tikz}
\usetikzlibrary{arrows.meta, positioning, fit, backgrounds, calc}

\usepackage{pifont}

\title{OLIVE: View-Augmented Latent Prediction with Waveform Reconstruction for Speech SSL}

\author{%
  Karl El Hajal\textsuperscript{1,2}, Mathew Magimai.-Doss\textsuperscript{1} \\
  \textsuperscript{1}Idiap Research Institute, Switzerland \\
  \textsuperscript{2}EPFL, Switzerland \\
  \texttt{\{karl.elhajal, mathew\}@idiap.ch}
}

\begin{document}

\maketitle

\begin{abstract}
    We propose \textbf{O}nline \textbf{L}atent prediction with \textbf{I}nvariant \textbf{V}iews and r\textbf{E}construction (OLIVE), a self-supervised speech representation learning framework that jointly optimizes analysis and synthesis objectives. OLIVE couples masked latent prediction and waveform reconstruction through a functional separation within a single encoder: reconstruction constrains early encoder features to retain signal-level information, while masked latent prediction shapes later contextual representations for robust downstream performance, with waveform view augmentations defining the invariances they learn. To the best of our knowledge, OLIVE is the first speech SSL model that directly reconstructs waveforms, retains its vocoder after pre-training, and remains competitive on discriminative tasks. Adding reconstruction largely preserves the speech recognition performance of masked latent prediction alone, while improving waveform reconstruction and speaker tasks. Adding waveform mixup and gain augmentations provides a controllable way to shift the representations further toward speaker, generation, and reconstruction tasks, while remaining competitive on recognition and semantic tasks.
\end{abstract}

\section{Introduction}

Recent advances in speech representation learning have been largely driven by
self-supervised learning (SSL) on large-scale unlabeled audio \citep{mohamed2022self, liu2022audio}. Models such as
wav2vec~2.0~\citep{baevski2020wav2vec2}, HuBERT~\citep{hsu2021hubert},
WavLM~\citep{chen2022wavlm}, and data2vec~\citep{baevski2022data2vec,baevski2022data2vec2}
learn contextual representations that serve as general-purpose
feature extractors, supporting both linguistic tasks such as automatic speech
recognition~\citep{baevski2020wav2vec2,hsu2021hubert}
and paralinguistic tasks including speaker and other non-linguistic acoustic
attributes such as diarization and speech emotion recognition~\citep{yang2021superb,baroudi2024specializing,atmaja2022evaluating,purohit2025emotion}.
These methods inherit ideas from contrastive prediction~\citep{oord2018representation},
masked modeling~\citep{devlin2019bert,liu2020mockingjay}, quantization and
discrete target discovery~\citep{ling2020decoar,liu2020non}, and
teacher--student distillation~\citep{grill2020byol,caron2021emerging}.
Existing approaches are nevertheless predominantly focused on
discriminative tasks: they aim to extract representations that are maximally
informative for downstream predictive tasks, without explicitly preserving the
information required for speech signal generation.

In contrast, speech communication fundamentally arises from two tightly coupled
phenomena: \emph{perception} (hearing) and \emph{production} (speaking).
Classical speech processing has long embraced this duality through the
\emph{analysis--synthesis} paradigm, where signals are decomposed into
structured components and subsequently reconstructed~\citep{atal1971speech, mcaulay1986speech, ghitza1986speech}. This framework leverages the fact
that speech production is governed by a well-defined physical and physiological
process, making it amenable to explicit modeling~\citep{fant1960acoustic, titze1994principles, zhang2016mechanics}. Recent neural work has revisited this perspective by training decoders on frozen self-supervised representations, effectively forming a neural analysis--synthesis pipeline~\citep{choi2021neural, polyak2021speech, siuzdak2022wavthruvec, guo2024vec2wav}.
This has enabled a range of synthesis approaches for tasks such as voice conversion~\citep{baas2023voice}, text-to-speech~\citep{el2025knn}, and speech enhancement~\citep{irvin2023self} that leverage the structured information encoded in intermediate self-supervised representations.

Despite this motivation, synthesis remains peripheral in most speech SSL
systems, and coupling analysis and synthesis during pre-training is non-trivial. Reconstruction requires signal-level information such as speaker
characteristics and fine-grained acoustic detail, whereas discriminative SSL benefits from
representations that abstract away such variability, so a
reconstruction objective risks reducing the discriminative quality of the contextual representations that
make SSL effective for recognition. UniWav, which trains an encoder jointly with a generative decoder,
reports lower recognition performance than SSL models of similar
scale as its main limitation~\citep{liu2025uniwav}. 
data2vec-SG keeps recognition close to data2vec, but treats reconstruction as a
lightly weighted auxiliary loss on its top contextual layers, and its decoder is discarded after pre-training~\citep{wang2023data2vec}. 
It thus remains open how to
reconstruct waveforms directly, retain a functional decoder, and preserve the discriminative strength of speech SSL.

In this work, we propose \textbf{OLIVE} (\textbf{O}nline \textbf{L}atent prediction with \textbf{I}nvariant \textbf{V}iews and r\textbf{E}construction), a self-supervised speech representation learning framework that jointly optimizes analysis and synthesis objectives. OLIVE resolves this tension through a \emph{functional separation} within a single encoder: the synthesis branch constrains early local encoder features to preserve information for waveform reconstruction, while the analysis branch, based on masked latent prediction, shapes later contextual representations toward invariance
for robust downstream analysis. This stands in contrast to purely discriminative models, which may
discard information relevant to generation, and purely generative approaches,
which may underperform on discriminative tasks. Within the analysis branch,
independently augmented waveform views for the student and teacher act as an
explicit prior over the learned invariances, which can be adapted to the
requirements of different applications. We validate OLIVE against established SSL baselines on a broad suite of
downstream tasks and on direct waveform reconstruction, and study both the
effect of joint training and the effect of waveform view augmentations. Code and model weights will be released upon publication.

\paragraph{Contributions.}
\begin{itemize}[leftmargin=1.5em,itemsep=1pt,topsep=2pt,parsep=0pt]
    \item We introduce \textbf{OLIVE}, a speech SSL framework that jointly optimizes analysis and synthesis in a single pre-training stage; to our knowledge, it is the first speech SSL model that directly reconstructs waveforms, retains its vocoder after pre-training, and remains competitive on discriminative tasks.
    \item We propose a functional separation in which waveform reconstruction acts on early local encoder features, while later contextual representations are shaped primarily by masked latent prediction.
    \item We show that adding reconstruction largely preserves the recognition performance of masked latent prediction alone, while improving reconstructability, yielding a perceptually effective integrated decoder, and restoring speaker information.
    \item We find that waveform view augmentations provide a controllable way to shift representations toward speaker, generation, and reconstruction tasks while remaining competitive on recognition and semantic tasks.
\end{itemize}

\section{Related Work}

\paragraph{Self-supervised speech representation learning.}
Early speech SSL methods learned from raw waveforms using contrastive objectives that
predict latent representations of future or masked inputs from
past context~\citep{oord2018representation,schneider2019wav2vec,van2017neural,baevski2019vq,chung19_interspeech}.
Wav2vec~2.0 introduced masked latent prediction with quantized targets and a
contrastive loss at masked time steps~\citep{baevski2020wav2vec2}. HuBERT
replaced online quantization with masked prediction of offline clustered
units~\citep{hsu2021hubert}, while WavLM extended this line with denoising and
large-scale pre-training for a broader set of speech tasks~\citep{chen2022wavlm}.
Data2vec instead predicts continuous contextualized teacher representations, avoiding
discrete targets~\citep{baevski2022data2vec}.
Data2vec~2.0 improves the efficiency of this masked self-distillation setup by
reusing teacher targets across masked views, encoding only unmasked steps, and
using a lightweight convolutional decoder~\citep{baevski2022data2vec2}. OLIVE
uses masked latent prediction within a broader analysis--synthesis framework:
the student and teacher receive independently augmented waveform views, and a
synthesis branch performs waveform reconstruction from local encoder features.

\paragraph{View-based self-distillation.}
View-based self-supervised learning trains representations to agree across
stochastic views of the same input~\citep{schmidhuber1993predictable}.
BYOL showed that an online network can predict the representation of a slowly
updated target network without explicit negative pairs, learning invariances
through data augmentations~\citep{grill2020byol}.
In audio, BYOL-A adapts this principle using mixup and time--frequency transformations
on log-mel spectrograms~\citep{niizumi2021byol}, and BYOL-S further studies
bootstrapped speech representations across encoder architectures and
pre-training data choices~\citep{elbanna2022byols}. WaveBYOL extends this approach
to raw waveforms using time-domain augmentations, primarily targeting general
audio tasks~\citep{kim2023wavebyol}.
Similar to data2vec, WavJEPA predicts latent targets from raw waveforms~\citep{yuksel2025wavjepa}, following the broader
JEPA framework~\citep{assran2023self}. OLIVE builds on this
line by combining masked latent prediction with
time-domain waveform augmentations to explicitly induce invariances, while jointly
optimizing a synthesis objective.

\paragraph{Speech reconstruction and analysis--synthesis.}
Recent work trains neural vocoders \citep{oord2016wavenet, prenger2019waveglow, kumar2019melgan, yamamoto2020parallel} such as HiFi-GAN~\citep{kong2020hifigan} on frozen self-supervised features, effectively forming
neural analysis--synthesis pipelines~\citep{choi2021neural,polyak2021speech,siuzdak2022wavthruvec,guo2024vec2wav}, enabling synthesis methods that leverage self-supervised representations. Such methods typically treat synthesis
as a downstream process, rather than integrating it into
representation learning during pre-training. Neural audio codecs further explore this analysis--synthesis
interface by learning compact latent representations that can be decoded back
to waveforms~\citep{zeghidour2021soundstream,defossez2022high,liu2024codecresynthesis,casanova2025nanocodec}. These
works primarily target efficient storage, transmission, or speech language model inference, whereas OLIVE
uses reconstruction as a training objective for representation learning.

Closer to speech SSL pre-training, data2vec-SG adds a decoder that reconstructs
the clean waveform from the top contextual layers, and discards it after
pre-training~\citep{wang2023data2vec}. UniWav conditions a flow-matching decoder
on a weighted sum of all encoder layers to generate intermediate codec
latents rather than waveforms, but reports lower recognition performance than
SSL models of similar scale as its main limitation~\citep{liu2025uniwav}. Related two-stage
approaches first learn latent representations and then train a separate
reconstruction pathway~\citep{ioannides2025jepatokenizer}, while masked
generative pre-training methods such as Metis operate on SSL features for
downstream speech generation~\citep{wang2025metis}. OLIVE differs along three
axes: its decoder is conditioned on early
local encoder features, leaving later contextual representations to be shaped
primarily by the masked distillation objective; it reconstructs the input
waveform directly; and it is retained as a usable vocoder after pre-training.
Both objectives are optimized in a single pre-training stage, and OLIVE remains
competitive with SSL models of similar scale on speech recognition
(Section~\ref{sec:superb}).

\begin{figure}[t]
\centering
\includegraphics[width=.9\linewidth]{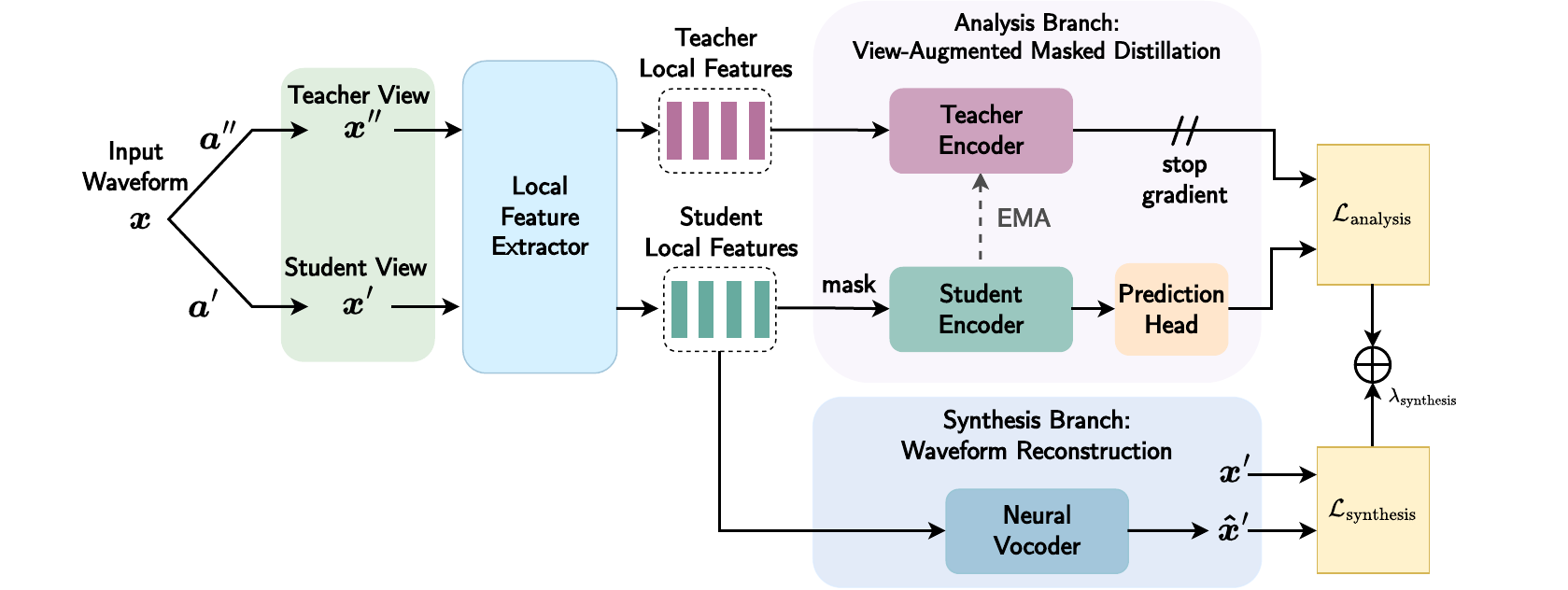}
\caption{
\textbf{OLIVE pre-training framework.} Two independently augmented waveform views are passed
through a shared local feature extractor. The analysis branch performs view-augmented masked
distillation: the student predicts contextual teacher targets from masked input features. The synthesis branch performs waveform reconstruction by conditioning a neural
vocoder on the local student features. The final objective combines the
analysis loss with a weighted synthesis loss.
}
\label{fig:olive_framework}
\end{figure}

\section{Method}

OLIVE learns speech representations by jointly solving two self-supervised
tasks, which we term \emph{analysis} and \emph{synthesis}. The framework is
illustrated in Figure~\ref{fig:olive_framework}. The analysis task encourages the
encoder to produce contextual representations that are invariant to
augmentation-induced variability, while the synthesis task encourages early local
encoder features to preserve acoustic information relevant to waveform reconstruction.
We refer to this design as a \emph{functional separation}: the two
objectives share one encoder but act at different depths, so
reconstruction can add signal-level information without directly constraining
the contextual representations used for recognition.

\subsection{Analysis: View-Augmented Masked Distillation}
\label{sec:analysis}

The analysis branch uses the teacher--student masked prediction objective of
data2vec~2.0~\citep{baevski2022data2vec2} with the addition of independently sampled waveform
views for the student and teacher. The augmentation family acts as an explicit prior over the learned invariances
(e.g., to environment, speaker, pitch, time, or gain) and can be adapted to applications with different requirements. We use $'$ to denote student views and $''$ to denote teacher
views. Given an input waveform $x$, we sample two stochastic augmentations $a'$ and $a''$ and form the student and teacher views
\begin{equation}
x' = a'(x), \qquad x'' = a''(x).
\end{equation}
The student processes a masked view of $x'$, while the teacher processes the
unmasked view $x''$. The student is trained to predict contextual latent targets
provided by the teacher at masked positions.
We decompose the student and teacher encoders into a local feature extractor
followed by a contextual encoder, and masking is applied to local latent
features before contextualization.

Let $e_{\theta}$ denote the student local feature extractor with parameters
$\theta$, $f_{\theta}$ the student contextual encoder, and
$e_{\bar{\theta}}, f_{\bar{\theta}}$ their teacher counterparts. Let
$m(\cdot)$ denote the masking operation applied in latent space to the student
local features. The student and teacher local feature sequences over $T$ frames are
\begin{equation}
r^{\prime}_{1:T} = e_{\theta}(x'),\qquad
\bar{r}^{\prime\prime}_{1:T} = e_{\bar{\theta}}(x''),
\end{equation}
and the corresponding contextual representations are
\begin{equation}
h^{\prime}_{1:T} = f_{\theta}(m(r^{\prime}_{1:T})),\qquad
\bar{h}^{\prime\prime}_{1:T} = f_{\bar{\theta}}(\bar{r}^{\prime\prime}_{1:T}).
\end{equation}
The teacher parameters are updated as an exponential moving average (EMA) of the student
parameters:
\begin{equation}
\bar{\theta} \leftarrow \tau \bar{\theta} + (1-\tau)\theta,
\end{equation}
where $\tau \in [0,1]$ is the target decay rate.

Teacher targets average the instance-normalized ($\operatorname{IN}$) outputs of the top $K$
teacher encoder layers. With $\bar{h}^{\prime\prime,\ell}_t$ the
teacher representation at layer $\ell \in \{1,\dots,L\}$ and time step $t$,
where $L$ is the number of encoder layers, the target is
\begin{equation}
z^{\prime\prime}_t
=
\frac{1}{K}
\sum_{\ell=L-K+1}^{L}
\operatorname{IN}\!\left(\bar{h}^{\prime\prime,\ell}_t\right).
\end{equation}

The student predicts these targets at masked time steps using a mean-squared error objective:
\begin{equation}
\mathcal{L}_{\mathrm{analysis}}
=
\frac{1}{|\mathcal{M}|}
\sum_{t \in \mathcal{M}}
\left\|
p_{\theta}(h^{\prime}_t)
-
\mathrm{sg}(z^{\prime\prime}_t)
\right\|_2^2
\cdot d^{-1/2},
\end{equation}
where $\mathcal{M}$ is the set of masked frames, $p_{\theta}$ is a prediction
head, $d$ is the representation dimension, and $\mathrm{sg}(\cdot)$ denotes stop-gradient.

\subsection{Synthesis: Waveform Reconstruction}

The synthesis branch reconstructs the waveform from early encoder features with
a HiFi-GAN vocoder~\citep{kong2020hifigan}. HiFi-GAN is an adversarially trained generator
paired with waveform discriminators that distinguish real from generated audio.
We replace the usual mel-spectrogram conditioning with the OLIVE encoder
representation, so that reconstruction directly constrains the representation
itself. Using the unmasked student local feature sequence $r^{\prime}_{1:T}$ extracted from the
waveform view $x'$, the
generator $G_{\psi}$ with parameters $\psi$ produces a
waveform reconstruction $\hat{x}' = G_{\psi}(r^{\prime}_{1:T})$.

The adversarial component uses two discriminator families. The
multi-period discriminator (MPD) contains sub-discriminators with different
periods, capturing harmonic and pitch-period structure. The
multi-scale discriminator (MSD) operates at several temporal resolutions,
encouraging realistic structure across scales~\citep{kumar2019melgan,kong2020hifigan}. The synthesis objective combines
mel-spectrogram reconstruction, adversarial training, and feature matching.

The mel reconstruction loss is $\mathcal{L}_{\mathrm{mel}} = \|\operatorname{mel}(x')-\operatorname{mel}(\hat{x}')\|_{1}$,
where $\operatorname{mel}(\cdot)$ is the log-mel spectrogram transform. Let $\mathcal{D}_{\phi}=\{D_{\phi}^{j}\}_{j=1}^{J}$ be the
discriminators. The adversarial losses are:
\begin{equation}
\mathcal{L}_{\mathrm{gen}}
=
\sum_{j=1}^{J}
\left(D_{\phi}^{j}(\hat{x}')-1\right)^2,
\qquad
\mathcal{L}_{\mathrm{disc}}
=
\sum_{j=1}^{J}
\left[
\left(D_{\phi}^{j}(x')-1\right)^2
+
\left(D_{\phi}^{j}(\mathrm{sg}(\hat{x}'))\right)^2
\right].
\end{equation}

The feature-matching loss
$\mathcal{L}_{\mathrm{fm}} = \sum_{j=1}^{J}\sum_{i=1}^{T_j} N_{j,i}^{-1} \| D_{\phi,i}^{j}(x')-D_{\phi,i}^{j}(\hat{x}') \|_1$
compares the activations $D_{\phi,i}^{j}(\cdot)$ of layer $i$ in
sub-discriminator $j$, where $T_j$ is the number of feature layers and $N_{j,i}$
the number of elements in that activation. The mel-spectrogram term anchors the
reconstruction to the reference signal in a stable time-frequency space, while
the adversarial and feature-matching terms encourage perceptually realistic 
details, with $\lambda_{\mathrm{fm}}$ and $\lambda_{\mathrm{mel}}$
weighting the corresponding terms in the generator objective
$\mathcal{L}^{G}_{\mathrm{synthesis}} = \mathcal{L}_{\mathrm{gen}} + \lambda_{\mathrm{fm}}\mathcal{L}_{\mathrm{fm}} + \lambda_{\mathrm{mel}}\mathcal{L}_{\mathrm{mel}}$;
the discriminator objective is $\mathcal{L}^{D}_{\mathrm{synthesis}} = \mathcal{L}_{\mathrm{disc}}$.

\subsection{Joint Optimization}

The final objective balances invariant representation learning with waveform reconstruction,
$\mathcal{L}_{\mathrm{OLIVE}} = \mathcal{L}_{\mathrm{analysis}} + \lambda_{\mathrm{synthesis}} \mathcal{L}^{G}_{\mathrm{synthesis}}$,
where $\lambda_{\mathrm{synthesis}}$ controls the analysis--synthesis trade-off.
This objective updates the student encoder, prediction head, and generator,
while the teacher is updated by EMA and the discriminator parameters are
updated separately.

\section{Experiments}

We validate OLIVE in two ways. First, we evaluate learned representations on a
broad suite of downstream speech tasks. Second, we evaluate waveform
reconstruction quality. Section~\ref{sec:pretraining} presents implementation
and pre-training details, Section~\ref{sec:superb} presents the downstream
evaluation, and Section~\ref{sec:reconstruction_eval} presents the waveform
reconstruction results.

\subsection{Implementation and Pre-training Details}
\label{sec:pretraining}

We implement OLIVE in fairseq\footnote{\url{https://github.com/facebookresearch/fairseq}}.
The analysis branch implements independently augmented student and teacher views, and uses masked-prediction configurations from the data2vec~2.0
speech implementation. This is coupled with waveform reconstruction in the joint variants. We pre-train on the 960-hour LibriSpeech dataset~\citep{panayotov2015librispeech} and focus on \emph{Base}-sized models due to
computational constraints. We train and evaluate four variants:
\textbf{OLIVE-A (Mix)} is an analysis-only variant using waveform mixup augmentations, and
\textbf{OLIVE-A (Mix+Gain)} additionally applies gain perturbation. The joint
variants, denoted \textbf{OLIVE-J}, optimize both masked prediction and waveform
reconstruction:
\textbf{OLIVE-J (Mask-only)} does not use waveform augmentations, while \textbf{OLIVE-J (Mix)} uses waveform mixup. All four variants share the same 93.8M-parameter
analysis architecture; the OLIVE-J variants also share the same synthesis architecture and loss.
To select which analysis-only variants to train, we first trained
\emph{Small}-sized models with hidden size $d=384$ and evaluated their
performance on a representative subset of downstream tasks (Appendix~\ref{app:variant_selection_ablation}).
We describe the OLIVE pre-training setup in three parts: (i) the shared
encoder and masked distillation objective, (ii) the waveform view
augmentations used, and (iii) the synthesis branch in the joint models.

\paragraph{Encoder and masked distillation.}
OLIVE uses a raw-waveform speech encoder architecture following prior SSL
models~\citep{baevski2020wav2vec2,baevski2022data2vec2}, with a 7-layer
convolutional waveform encoder, a convolutional positional encoder, and a
12-layer Transformer~\citep{vaswani2017attention} with hidden size $d=768$.
The waveform encoder has an overall stride of 320 samples, corresponding to
20~ms at 16~kHz. It is shared between the student and teacher networks. Teacher targets are formed by instance-normalizing and
averaging the top $K=8$ teacher layers, and the student predicts these targets
with a convolutional prediction head.
Similarly to data2vec~2.0, we use
inverse-block masking and eight masked views per sample. Final models are
trained for 400k updates with Adam and cosine learning-rate decay. The teacher
EMA decay $\tau$ is annealed from $0.999$ to $0.99999$ over the first 75k updates. We use
mixed precision for the analysis branch and FP32 for the synthesis branch
for numerical stability. Further implementation and compute details are reported in Appendices~\ref{app:pretraining_details} and~\ref{app:compute_details}.

\paragraph{Waveform view augmentations.}
We use waveform-level augmentations to define invariances for the analysis
objective. We focus on perturbations related to acoustic environment and
recording conditions, while avoiding transformations such as pitch shifting or
time-scale modification that may remove information relevant to speaker
identity, prosody, and paralinguistic tasks.
\textbf{Mixup} mixes the input waveform with an utterance sampled from a memory
bank using linear interpolation with a mixing ratio uniformly sampled from
$[0, 0.4]$, encouraging
robustness to background acoustic interference and following the general
regularization principle of mixup-style training~\citep{zhang2018mixup,niizumi2021byol}. \textbf{Gain} applies a random global
gain followed by clamping to $[-1,1]$, encouraging invariance to recording gain
and mild clipping artifacts. We omit gain perturbation in joint training, since clipping
artifacts may adversely affect waveform reconstruction.

\paragraph{Synthesis branch.}
For OLIVE-J, the synthesis branch reconstructs the student view $x'$ from its
local encoder features with a HiFi-GAN V2 vocoder~\citep{kong2020hifigan}. Synthesis training uses
7040-sample waveform segments ($0.44$~s at 16~kHz) randomly sampled from the full extracted waveform. In the final OLIVE-J
models, we set $\lambda_{\mathrm{synthesis}}=25$, chosen after ablations
(Appendix~\ref{app:loss_weight_ablation}).
A key design choice is which encoder representation conditions the vocoder.
Earlier layers typically preserve the fine-grained acoustic detail needed for waveform reconstruction; prior work on synthesis-based speech enhancement found that vocoders conditioned on SSL local features can benefit from the acoustic detail available before deeper contextual abstraction~\citep{irvin2023self}.
Our ablations confirm this under matched separate-vocoder training
(Appendix~\ref{app:hifigan_layer_ablation}), so we condition the vocoder on
local encoder features.
OLIVE-J optimizes 165.6M parameters during training, of which 70.7M belong to
discriminators used only for adversarial training and discarded afterwards;
waveform reconstruction then uses only the shared CNN feature extractor (4.2M)
and the HiFi-GAN V2 generator (1.1M) (Appendix~\ref{app:param_breakdown}).

\subsection{Downstream Evaluation}
\label{sec:superb}

We use the Speech Processing Universal PERformance Benchmark (SUPERB)~\citep{yang2021superb,tsai2022superb} using the S3PRL
toolkit.\footnote{\url{https://github.com/s3prl/s3prl}}
SUPERB evaluates frozen SSL models as general-purpose feature extractors. For
each task, a learned weighted sum of all encoder layers is fed to a lightweight
task head, and only the layer weights and the head are trained on labeled data.

\paragraph{Tasks.}
We evaluate OLIVE across the five SUPERB task categories: content, speaker,
paralinguistic, semantic, and generation. The downstream tasks are phoneme
recognition (PR), automatic speech recognition (ASR), keyword spotting (KS),
query-by-example spoken term detection (QbE), speaker identification (SID),
automatic speaker verification (ASV), speaker diarization (SD), emotion
recognition (ER), intent classification (IC), slot filling (SF), speech
translation (ST), speech enhancement (SE), source separation (SS), and voice
conversion (VC). Unlike some prior work~\citep{chen2022wavlm}, we do not perform task-specific hyperparameter search due to compute constraints; all models are evaluated with the same default task
configurations in S3PRL.
\paragraph{Baselines.}
All SSL baselines in the main comparison are Base-sized models trained on
LibriSpeech 960h and have similar encoder architectures. Therefore, the
comparison primarily isolates differences in pre-training objectives rather
than data, architecture, or model capacity. We compare OLIVE with
handcrafted log mel-filterbank acoustic features (FBANK) and widely used
speech SSL baselines: wav2vec~2.0 \citep{baevski2020wav2vec2}, HuBERT
\citep{hsu2021hubert}, WavLM \citep{chen2022wavlm}, data2vec
\citep{baevski2022data2vec}, and data2vec~2.0
\citep{baevski2022data2vec2}. We use official public checkpoints and
implementations from S3PRL. For data2vec-SG~\citep{wang2023data2vec} and
UniWav~\citep{liu2025uniwav}, no public checkpoints are provided. For
completeness, we include paper-reported data2vec-SG results ($\dagger$) on the
overlapping generation tasks (see Appendix~\ref{app:d2vsg_relative} for differences relative to each evaluation's data2vec baseline). We do not include UniWav in the main comparison
because it differs substantially in model scale and pre-training data.

\paragraph{Scores.}
We report two aggregate scores. First, we use the SUPERB$_s$ metric~\citep{feng2023superb},
which measures the fraction of the gap between the FBANK baseline and a
task-specific reference score denoted as SOTA. For model $u$, task set $T$,
and metric set $I_t$ for task $t$, we compute
\[
\mathrm{SUPERB}_s(u)
=
\frac{1000}{|T|}
\sum_{t \in T}
\frac{1}{|I_t|}
\sum_{i \in I_t}
\frac{s_{t,i}(u)-s_{t,i}(\mathrm{FBANK})}
{s_{t,i}(\mathrm{SOTA})-s_{t,i}(\mathrm{FBANK})}.
\]
For tasks with multiple evaluation metrics, normalized metric scores are averaged within
the task before averaging across tasks.
We also report a WavLM-style aggregate score \citep{chen2022wavlm},
\[
\mathrm{WavLM}_s(u) = \frac{1}{|T|}\sum_{t \in T}\tilde{s}_t(u),
\]
which averages per-task utility scores $\tilde{s}_t$ after metric normalization: error rates such as WER and EER are converted to utilities using
$100-\mathrm{error}$, QbE is scaled by $100$, and higher-is-better metrics are used directly.

\paragraph{Main results.}
\input{tables/superb}

Task-level results and overall SUPERB$_s$ and WavLM$_s$ scores are presented in
Tables~\ref{tab:superb-main-results} and~\ref{tab:superb-main-results-gen}.
The OLIVE variants achieve strong aggregate results across downstream tasks,
with their different setups shifting the balance between recognition and tasks
tied to fine-grained acoustic detail.
Specifically,
\textbf{OLIVE-A (Mix+Gain)} achieves the best SUPERB$_s$ score, whereas
\textbf{OLIVE-J (Mix)} achieves the best WavLM$_s$ score and the strongest speaker
and generation category scores.
\textbf{OLIVE-J (Mask-only)} ties for the second-best WavLM$_s$ score,
achieves strong ASR results (4.9 WER versus 4.8 for data2vec~2.0; see
Appendix~\ref{app:asr_finetuning} for results across LibriSpeech splits), and
leads on SF.
Nonetheless, the augmented OLIVE variants remain competitive on ASR with wav2vec~2.0, HuBERT, and WavLM.
Appendix~\ref{app:superb_radials} illustrates these differences through
normalized per-task and category-level profiles.
Appendix~\ref{app:superb_layerwise} visualizes layer-combination
weights for a representative set of tasks, illustrating the emphasis on earlier
layers for speaker and generation tasks and on later layers for content and
semantic tasks. Overall, the functional separation retains strong recognition
performance and improves speaker identification, while waveform view augmentations further steer the
representations toward tasks that rely on acoustic detail and speaker
information.

\subsection{Waveform Reconstruction Evaluation}
\label{sec:reconstruction_eval}

\paragraph{Setup.}
We evaluate waveform reconstruction directly by training a separate
feature-conditioned HiFi-GAN V2 vocoder from scratch on the frozen features of
each evaluated model, and comparing them with the integrated vocoder retained
from each OLIVE-J variant. All separate vocoders use the same HiFi-GAN V2 architecture and training setup as the jointly trained vocoders. They are trained on the LibriSpeech training set with mixup augmentation
(Appendix~\ref{app:hifigan_augmentation_ablation}) for 400k updates
(Appendix~\ref{app:vocoder_training}) and evaluated on LibriSpeech
\texttt{test-clean}; results are thus matched comparisons under a common
configuration, not fully converged upper bounds.

\paragraph{Models.}
The reconstruction baselines include mel-spectrogram, wav2vec~2.0,
HuBERT, WavLM, data2vec, and data2vec~2.0 features. For SSL representations,
we condition the vocoder on local encoder features, which preserve the most
acoustic detail (Appendix~\ref{app:hifigan_layer_ablation};~\citealp{irvin2023self}) and match the feature level used by the OLIVE-J synthesis branch. Among the
analysis-only OLIVE variants, we report \textbf{OLIVE-A (Mix+Gain)} based on
the layer ablation in
Appendix~\ref{app:hifigan_layer_ablation}.
For each of \textbf{OLIVE-J (Mask-only)} and \textbf{OLIVE-J (Mix)}, we report
two reconstruction conditions. The \emph{integrated vocoder} is the generator
retained from joint pre-training. The \emph{frozen feature-conditioned}
vocoder is trained separately with the same HiFi-GAN V2 setup as the baselines
on frozen features from the jointly pre-trained encoder; this
isolates the reconstructability of the learned features.

\paragraph{Metrics.}
We report spectral, intrusive, and perceptual metrics. Spectral fidelity is
assessed using log-mel spectrogram L1 distance
(Mel-L1), log-spectral distance (LSD), and mel-cepstral distortion (MCD). Pitch fidelity
is measured with frame-level fundamental-frequency mean absolute error ($F_0$
MAE). We additionally report intrusive intelligibility and quality metrics,
namely STOI \citep{taal2011algorithm} and PESQ \citep{rix2001perceptual}, as well as scale-invariant signal-to-distortion ratio
(SI-SDR) and signal-to-noise ratio (SNR). We also include UTMOS \citep{saeki2022utmos} and ViSQOL \citep{hines2015visqol} as learned perceptual metrics.
Generated and reference signals are loudness-matched.

\input{tables/hifigan_compact}

\paragraph{Main results.}
Results are shown in Table~\ref{tab:resynthesis_main}.
OLIVE substantially improves waveform reconstructability. We first observe
that the analysis-only \textbf{OLIVE-A (Mix+Gain)} model, trained with
waveform augmentations, is competitive with or better than strong analysis-only baselines on most
reconstruction metrics. For the \textbf{OLIVE-J} models, vocoders conditioned on
frozen features outperform the analysis-only baselines across all metrics under
matched separate-vocoder training. \textbf{OLIVE-J (Mask-only)} with a frozen feature-conditioned vocoder
achieves the best Mel-L1, LSD, and UTMOS scores.
\textbf{OLIVE-J (Mix)} with a frozen feature-conditioned vocoder is strongest on
all other metrics. Meanwhile, the integrated \textbf{OLIVE-J} vocoders perform
strongly on perceptual and signal-level metrics while being less precise on
Mel-L1 and LSD, which likely reflects the demands of joint training on the vocoder. Overall, joint analysis--synthesis pre-training makes the learned representation more informative for reconstruction, while yielding a perceptually effective decoder.
Appendix~\ref{app:reconstruction_spectrograms} provides spectrogram
visualizations of reconstructed waveforms, and Appendix~\ref{app:resynthesis_wer_secs}
reports intelligibility (WER) and speaker similarity (SECS) of the resynthesized speech.

\section{Analysis of Augmentation and Reconstruction}
\label{sec:knob_analysis}

We analyze the effects of augmentation and reconstruction on the learned representations through a two-factor ablation: data2vec~2.0 uses neither,
\mbox{OLIVE-A (Mix)} adds augmentation, \mbox{OLIVE-J (Mask-only)} adds reconstruction, and
\mbox{OLIVE-J (Mix)} combines both.
Figure~\ref{fig:knob_pareto} shows the effect of each component and
Figure~\ref{fig:knob_category} the resulting SUPERB category profiles.

\begin{figure*}[t]
\centering
\resizebox{\textwidth}{!}{\input{figures/knob_pareto}}
\vspace{-8pt}
\caption{\textbf{Effects of waveform view augmentation and reconstruction.}
x-axis: LibriSpeech \texttt{test-other} WER (no LM; Appendix~\ref{app:asr_finetuning}). Left: mean Speaker and
Generation SUPERB score (normalized per task: FBANK${}=0$, best${}=1000$).
Right: PESQ of frozen feature-conditioned vocoders, with OLIVE-A (Mix+Gain)
replacing OLIVE-A (Mix). Upper left is better.}
\label{fig:knob_pareto}
\end{figure*}

\paragraph{Augmentation trades frame-level precision for invariance.}
Adding mixup has the same effect with or without reconstruction
(Figure~\ref{fig:knob_category}): speaker and generation scores increase, as
does reconstruction quality on perceptual and signal-level metrics, while content and semantic scores
decrease slightly. Further, predicting
across independently augmented views favors tasks that pool over segments or
utterances (SID, SD, KS) at the expense of the frame-level detail required by
PR, ASR, and QbE. Among
generation tasks, SE and SS gain most, as mixup views resemble the
interference these tasks remove. The SUPERB VC task behaves differently: being
any-to-one, it rewards features that discard source-speaker identity, so the
baselines with the lowest SID accuracy (data2vec, data2vec~2.0) obtain among the
best VC scores.

\begin{wrapfigure}{r}{0.38\linewidth}
\vspace{-14pt}
\centering
\input{figures/superb_radar_macros}
\resizebox{0.8\linewidth}{!}{\endlinechar=-1 \input{figures/superb_category_2x2}}
\caption{\textbf{SUPERB category scores:}
\textcolor{taskData2vec}{\textbf{data2vec~2.0}},
\textcolor{taskOliveMix}{\textbf{OLIVE-A (Mix)}} (+aug.),
\textcolor{taskOliveMask}{\textbf{OLIVE-J (Mask-only)}} (+recon.),
\textcolor{taskOliveJ}{\textbf{OLIVE-J (Mix)}} (both), normalized as in Figure~\ref{fig:knob_pareto}.}
\label{fig:knob_category}
\vspace{-10pt}
\end{wrapfigure}

\paragraph{Reconstruction restores the acoustic detail discarded by masked latent prediction.}
Masked latent prediction regresses averaged, normalized contextual targets and
thus has no incentive to retain signal-level detail: the data2vec models have
the lowest baseline SID accuracy (Table~\ref{tab:superb-main-results})
and reconstruct waveforms no better than other SSL features
(Table~\ref{tab:resynthesis_main}). Adding reconstruction
improves every reconstruction metric in both settings (without augmentation,
PESQ from 2.21 to 2.81 and SI-SDR from $-9.7$ to $+1.0$~dB) and SID accuracy
(45.5 to 64.6 without augmentation, 66.9 to 83.1 with it), while PR and ASR are
nearly unchanged. This matches OLIVE's functional separation: reconstruction
acts on the local encoder features while masked prediction governs the
contextual layers, so acoustic detail is recovered without the recognition
cost of augmentation.

\paragraph{The two components are complementary.}
Combined with augmentation, reconstruction further improves generation, and partly offsets the content cost of
augmentation. Both OLIVE-J models therefore lie on the Pareto fronts of
Figure~\ref{fig:knob_pareto}: OLIVE-J (Mask-only) nearly matches data2vec~2.0 on
recognition with much better reconstruction, and OLIVE-J (Mix) obtains the
best speaker and generation scores and PESQ.
In practice, reconstruction suffices for recognition-oriented applications, while
speaker- and generation-oriented
applications benefit from combining it with mixup augmentations.

\section{Conclusion}

We introduced \textbf{OLIVE}, a speech SSL framework that jointly optimizes
analysis and synthesis during pre-training through a functional separation:
reconstruction acts on early local encoder features, while later contextual
representations are shaped by view-augmented masked latent prediction. Joint
training improves speaker, generation, and waveform reconstruction performance
over analysis-focused SSL representations, retains strong
content and semantic performance, and yields a perceptually effective
integrated decoder. In addition, waveform view augmentations offer a controllable
way to adapt the learned invariances to different applications. OLIVE's framework combines analysis and synthesis components in a modular fashion such that future advances in
representation learning and waveform synthesis can be readily incorporated.
Overall, this design enables a single pre-trained model to retain more acoustic
detail for synthesis while maintaining strong discriminative representations.

\paragraph{Limitations.}
Our study is limited to English
speech and Base-sized training regimes on LibriSpeech. Future work
should examine different model sizes, more diverse pre-training data, and
task-specific waveform augmentations.

\begin{ack}
This work was partially supported by the Swiss National Science Foundation through the project ``Pathological Speech Synthesis (PaSS)'', grant agreement no. 219726.
\end{ack}

\small
\bibliographystyle{unsrtnat}
\bibliography{references}

\clearpage
\appendix

\section{Model and Pre-training Details}
\label{app:pretraining_details}

Table~\ref{tab:model_pretraining_details} consolidates the main optimization,
encoder, and synthesis-branch configurations used for OLIVE pre-training.
\input{tables/model_pretraining_details}

\subsection*{Parameter Breakdown}
\label{app:param_breakdown}

Table~\ref{tab:parameter_breakdown} reports the parameter counts for the shared
encoder and synthesis components.
\input{tables/parameter_breakdown}

\subsection*{Separate Vocoder Training}
\label{app:vocoder_training}

Separate feature-conditioned HiFi-GAN V2 vocoders (Section~\ref{sec:reconstruction_eval})
are trained for 400k updates with batch size 72, Adam ($\beta_1=0.8$,
$\beta_2=0.99$), learning rate $7.5\times 10^{-4}$, and exponential
learning-rate decay of 0.999. Training uses 7040-sample waveform segments
($0.44$~s at 16~kHz), 80-bin log-mel spectrograms, a 1024-point FFT,
1024-sample window, and 320-sample hop size. Training data are augmented with
mixup, following the augmentation ablation in
Appendix~\ref{app:hifigan_augmentation_ablation}.

\section{Relative Comparison with data2vec-SG}
\label{app:d2vsg_relative}

The data2vec-SG results in Table~\ref{tab:superb-main-results-gen} are taken
from \citet{wang2023data2vec}. Baseline scores differ between the reported results and our evaluation, so absolute scores are not directly comparable. We
therefore provide a relative comparison: Table~\ref{tab:d2vsg_relative} reports
each model's difference from data2vec Base evaluated in the same setup.
\input{tables/d2vsg_relative}

\section{Ablations}
\label{app:ablations}

This section presents ablations used to determine pre-training configuration choices.

\subsection{Joint Loss Weight}
\label{app:loss_weight_ablation}

To choose the synthesis loss weight for OLIVE-J, we trained 100k-step joint
models with $\lambda_{\mathrm{synthesis}}\in\{1,25,50\}$ and evaluated them on a representative subset of
downstream discriminative tasks and waveform reconstruction quality on the LibriSpeech \texttt{dev-clean} set. A non-unit synthesis weight is needed in practice because the analysis
and synthesis objectives operate on different numerical scales. The analysis
branch implements the masked regression term from Section~\ref{sec:analysis} as
a framewise feature loss over the masked set $\mathcal{M}$:
\begin{equation}
\mathcal{L}_{\mathrm{analysis}}
=
\frac{1}{|\mathcal{M}|}
\sum_{t \in \mathcal{M}}
\sum_{k=1}^{d}
\frac{
\left(
p_{\theta}(h^{\prime}_t)_k
-
\mathrm{sg}(z^{\prime\prime}_t)_k
\right)^2
}{\sqrt{d}} .
\label{eq:appendix_analysis_scale}
\end{equation}
Although the factor $1/\sqrt{d}$ stabilizes the magnitude of the regression
term, the loss still accumulates errors over the $d$ latent dimensions before
averaging over masked frames, so moderate per-dimension errors can produce
loss values on the order of several units. By contrast, the HiFi-GAN terms in $\mathcal{L}^{G}_{\mathrm{synthesis}}$
are scalar objectives combined into the same joint loss. Since the analysis
loss accumulates over $d$ dimensions and masked frames, the two objectives operate at different numerical scales, which motivates a
non-unit synthesis weight. Tables~\ref{tab:loss_weight_subset} and
\ref{tab:loss_weight_reconstruction} (visualized in Figure~\ref{fig:loss_weight_pareto}) show that $\lambda_{\mathrm{synthesis}}=25$
provides the best balance: increasing the weight from 1 to 25 substantially
improves reconstruction, while increasing it further to 50 weakens several
downstream metrics and yields less consistent reconstruction gains.

\input{tables/loss_weight_ablation}

\begin{figure}[!ht]
\centering
\resizebox{0.9\linewidth}{!}{\input{figures/loss_weight_pareto}}
\caption{\textbf{Synthesis loss weight trade-off.} PR PER versus PESQ (left)
and MCD (right) of the integrated vocoder for $\lambda_{\mathrm{synthesis}}\in\{1,25,50\}$.}
\label{fig:loss_weight_pareto}
\end{figure}

\subsection{Variant Selection}
\label{app:variant_selection_ablation}

To select which analysis-only variants to train, we first trained
\emph{Small}-sized models with hidden size $d=384$ for 100k updates and evaluated
their performance on a representative subset of content, speaker,
paralinguistic, generation, and reconstruction tasks.
Table~\ref{tab:variant_selection_ablation} reports the resulting ablation
used to select which variants to train at full \emph{Base} scale. The final column reports a normalized aggregate score across all metrics,
under which Mix ranks first and Mix+Gain second among the OLIVE analysis-only
variants.

\input{tables/variant_selection_ablation}

\subsection{HiFi-GAN Layer Ablation}
\label{app:hifigan_layer_ablation}

This section reports how waveform reconstruction changes with the choice of
conditioning layer.
Table~\ref{tab:hifigan_layer_ablation} compares separate HiFi-GAN V2 vocoders
trained on frozen features from \textbf{OLIVE-A (Mix)}, \textbf{OLIVE-A (Mix+Gain)}, and \textbf{WavLM Base} at different encoder depths. In all cases, the
best reconstruction quality is obtained from local encoder features, with
quality degrading as conditioning moves deeper into the contextual encoder.
Among the OLIVE analysis-only variants at the local encoder level,
\textbf{OLIVE-A (Mix+Gain)} is
slightly stronger on perceptual metrics and pitch accuracy. It is
therefore selected for the final waveform reconstruction comparison.

\input{tables/hifigan_layer_ablation}

\subsection{HiFi-GAN Augmentation Ablation}
\label{app:hifigan_augmentation_ablation}

Table~\ref{tab:hifigan_augmentation_ablation} compares matched separate
HiFi-GAN V2 vocoders trained on the same frozen \textbf{OLIVE-A (Mix+Gain)}
local encoder features, with and without waveform mixup augmentation applied during training. Mixup improves all reported reconstruction metrics, so
we use it for all separate vocoder training in the main evaluation.

\input{tables/hifigan_augmentation_ablation}

\section{SUPERB Layerwise Profiles}
\label{app:superb_layerwise}

\begin{figure}[H]
\centering
\includegraphics[width=0.49\linewidth]{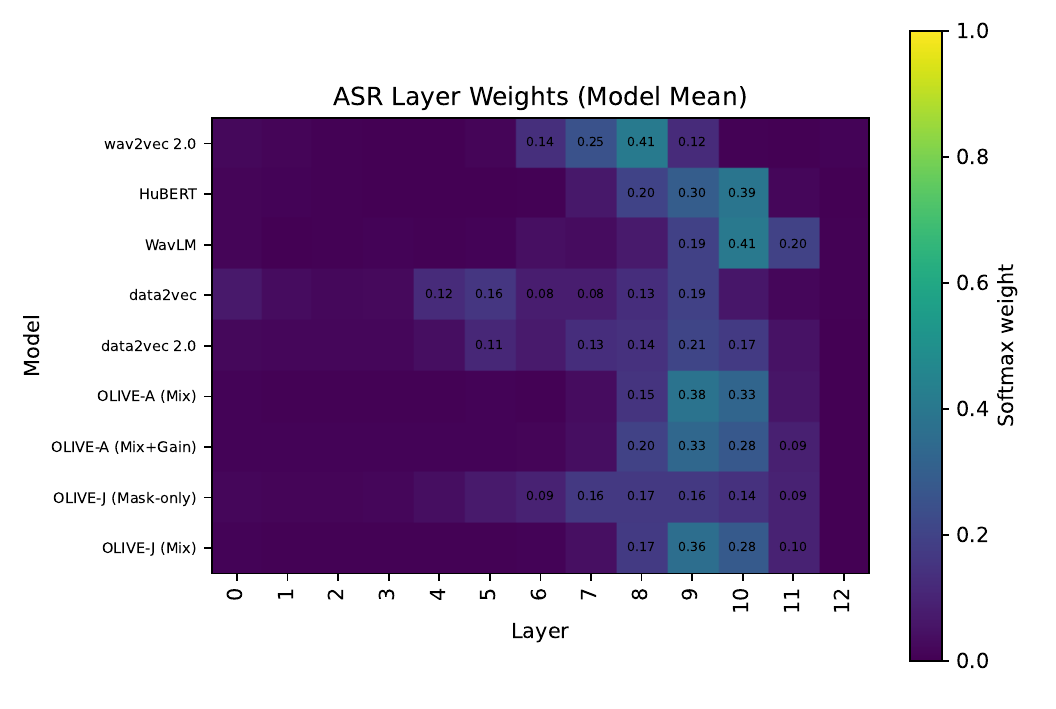}\hfill
\includegraphics[width=0.49\linewidth]{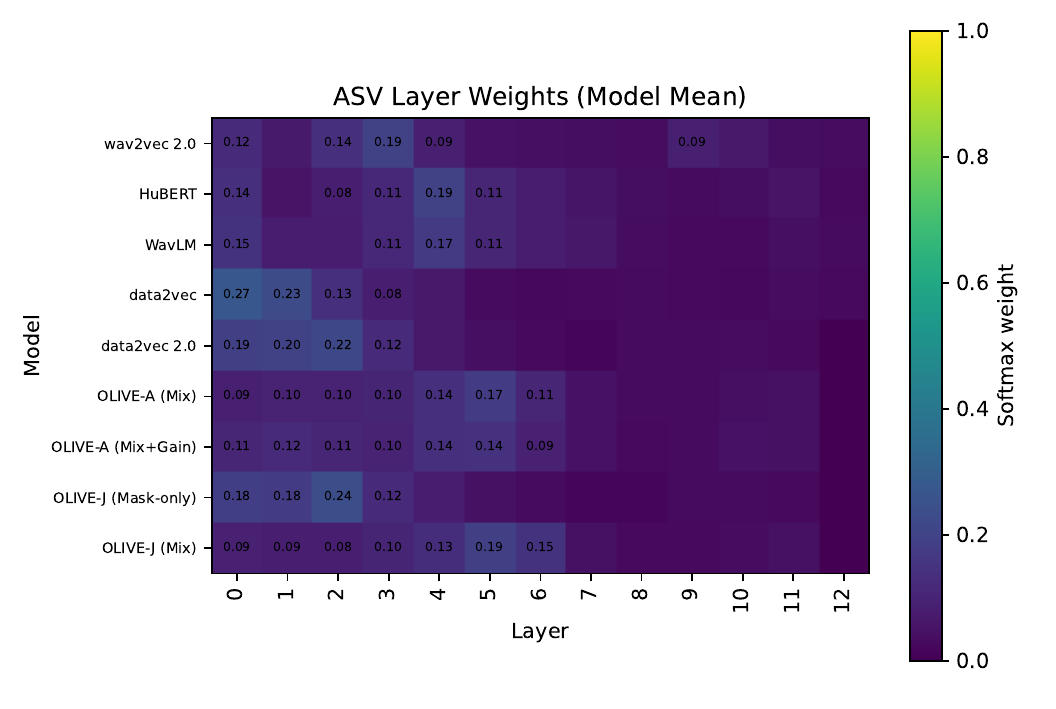}\\[4pt]
\includegraphics[width=0.49\linewidth]{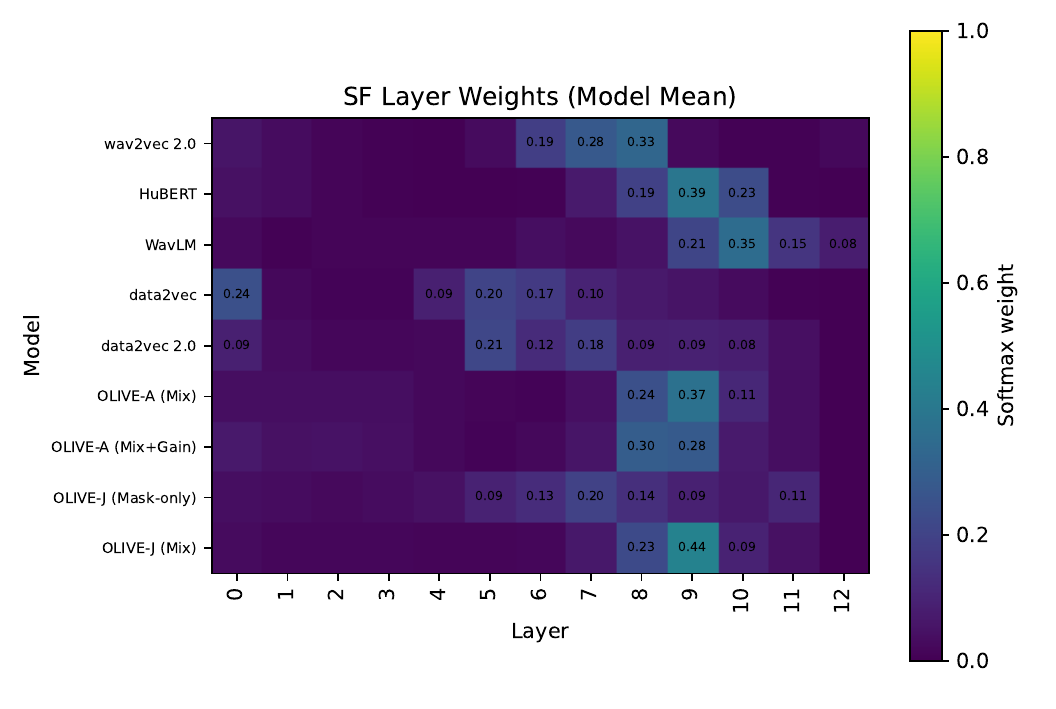}\hfill
\includegraphics[width=0.49\linewidth]{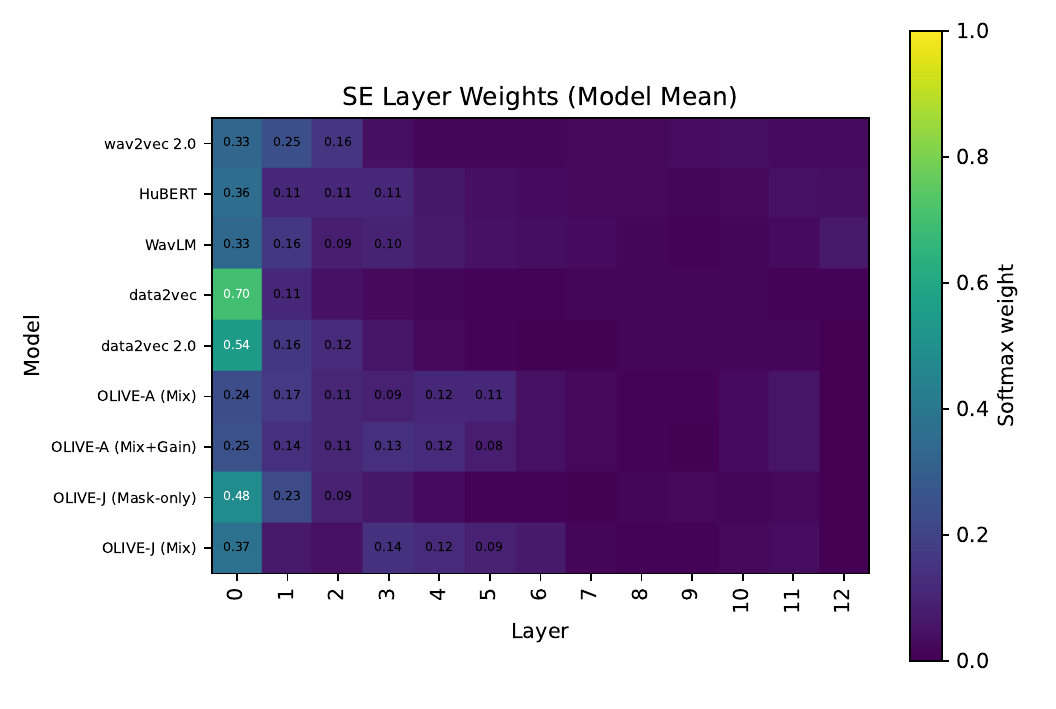}
\caption{\textbf{Representative SUPERB layer-combination weights.} Automatic
speech recognition (ASR) and slot filling (SF) emphasize later contextual
layers, while automatic speaker verification (ASV) and speech enhancement (SE)
place more weight on earlier layers.}
\label{fig:superb_layerwise}
\end{figure}

Figure~\ref{fig:superb_layerwise} shows representative learned layer
combination weights for four SUPERB tasks. ASR and slot filling place most of
their weight on later contextual layers, whereas speaker verification and
speech enhancement rely more heavily on earlier layers.

\section{SUPERB Radar Profiles}
\label{app:superb_radials}

Figure~\ref{fig:superb_task_radial} shows normalized per-task SUPERB profiles
for the OLIVE variants and the SSL baselines, together with the category-level
comparison.

\begin{figure*}[!ht]
\centering
\resizebox{\textwidth}{!}{
\input{figures/superb_three_radials}
}
\caption{
\textbf{SUPERB radar comparisons.} Left and center: normalized per-task
profiles. Right: aggregate category-level comparison. Per-task metrics are normalized so that FBANK is 0 and the best value in
Tables~\ref{tab:superb-main-results} and~\ref{tab:superb-main-results-gen} is
1000; multi-metric tasks are averaged within task, and category scores average
the tasks within each category.
}
\label{fig:superb_task_radial}
\end{figure*}

\section{ASR Fine-tuning Evaluation}
\label{app:asr_finetuning}

\input{tables/asr_lm_full}

We complement the SUPERB ASR evaluation by reporting word error rate on the
standard LibriSpeech development and test splits (\texttt{dev-clean},
\texttt{dev-other}, \texttt{test-clean}, and \texttt{test-other}), using
\texttt{train-clean-100} for fine-tuning under the same default task settings,
with and without a 4-gram language model; the results are shown in
Table~\ref{tab:asr-lm-full}. The strongest recognition numbers are
obtained by data2vec~2.0, with \textbf{OLIVE-J (Mask-only)} close behind on every
split. This suggests that the reconstruction objective, which acts mainly on
earlier encoder features, can enhance the information retained for reconstruction without degrading the discriminative quality of later contextual representations, while maintaining strong downstream performance. The OLIVE models trained with augmented waveform views show a recognition trade-off compared to data2vec~2.0 and OLIVE-J (Mask-only), but remain competitive with other
Base-sized SSL systems. \textbf{OLIVE-A (Mix+Gain)}, for example, stays close to
WavLM Base on most ASR splits while performing better on several speaker and
generation metrics. It achieves the highest overall SUPERB$_s$ score, while
\textbf{OLIVE-J (Mix)} achieves the highest overall WavLM$_s$ score.

\section{Intelligibility and Speaker Preservation of Resynthesized Speech}
\label{app:resynthesis_wer_secs}

We complement the signal-level evaluation with measures of linguistic content
and speaker identity preservation in Table~\ref{tab:resynthesis_wer_secs}.
All SSL-conditioned vocoders yield similar intelligibility, with the frozen feature-conditioned OLIVE-J vocoders
obtaining the lowest WER, while speaker similarity separates the models more
clearly: every OLIVE variant exceeds all SSL
baselines in SECS, and OLIVE-J (Mix) obtains the highest SECS with both the
frozen feature-conditioned and the integrated vocoder. Waveform view
augmentations further increase SECS over OLIVE-J (Mask-only), consistent with the speaker-task gains
observed on SUPERB.

\input{tables/resynthesis_wer_secs}

\section{Waveform Reconstruction Confidence Intervals}
\label{app:reconstruction_ci}

Tables~\ref{tab:resynthesis_test_clean} and~\ref{tab:resynthesis_test_clean_waveform}
report the waveform reconstruction results of Table~\ref{tab:resynthesis_main}
together with the margins of error from 95\% bootstrap confidence intervals.

\input{tables/hifigan}

\section{Waveform Reconstruction Spectrograms}
\label{app:reconstruction_spectrograms}

\begin{figure}[H]
\centering
\includegraphics[width=0.86\linewidth]{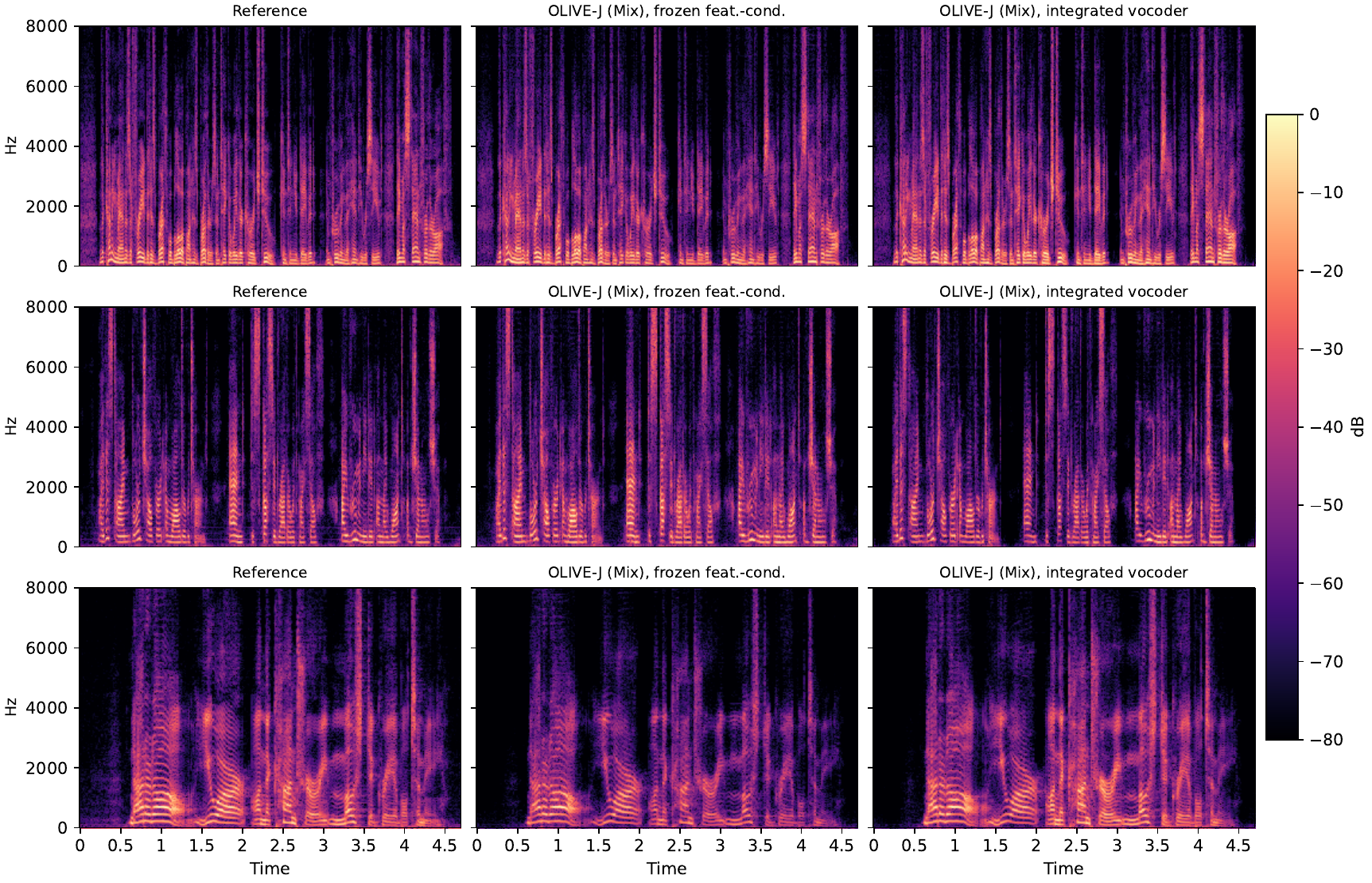}
\caption{Spectrograms for three reference utterances and reconstructions from
the frozen feature-conditioned and integrated OLIVE-J (Mix) vocoders.}
\label{fig:waveform_reconstruction_olive_j_spectrograms}
\end{figure}

\begin{figure}[H]
\centering
\includegraphics[width=\linewidth]{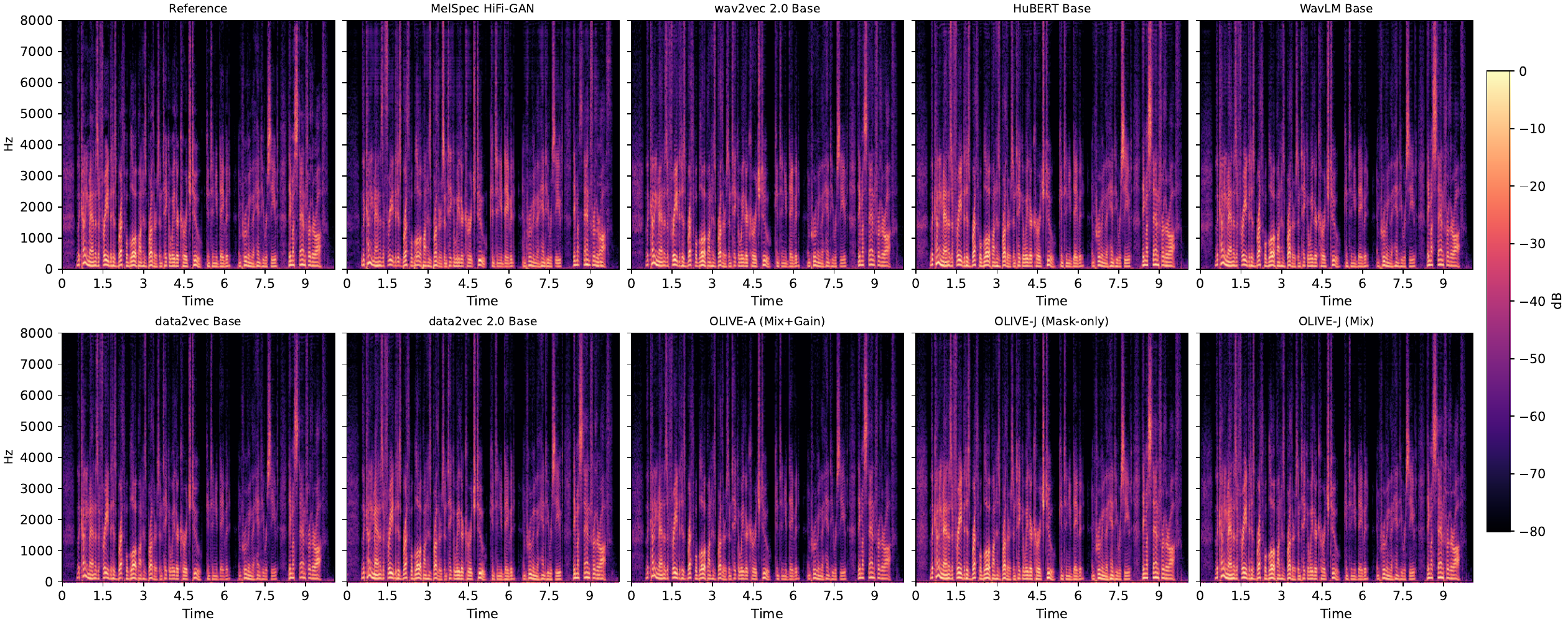}
\caption{Spectrograms for one reference utterance and reconstructions from
matched HiFi-GAN V2 vocoders conditioned on local features from the baseline and OLIVE
models.}
\label{fig:waveform_reconstruction_all_models_spectrograms}
\end{figure}

Figures~\ref{fig:waveform_reconstruction_olive_j_spectrograms} and
\ref{fig:waveform_reconstruction_all_models_spectrograms} show spectrograms that
complement Table~\ref{tab:resynthesis_main}.
Figure~\ref{fig:waveform_reconstruction_olive_j_spectrograms} illustrates three
reference utterances reconstructed by the frozen feature-conditioned and integrated
OLIVE-J (Mix) vocoders.
Figure~\ref{fig:waveform_reconstruction_all_models_spectrograms} illustrates one
reference utterance reconstructed by all separately trained, matched HiFi-GAN V2 vocoders conditioned on local encoder features from the baseline and OLIVE models.

\section{Compute Details}
\label{app:compute_details}

All main OLIVE pre-training runs used 2 NVIDIA H100 80GB GPUs. The final
analysis-only models required 123 training hours per run, and the final
OLIVE-J models required 206 training hours per run. Vocoder training and downstream task
evaluation used Multi-Instance GPU (MIG) 20GB and 40GB partitions of H100
GPUs. 

\end{document}

%% file: tables/superb.tex
\begin{table*}[t]
\centering
\footnotesize
\setlength{\tabcolsep}{4pt}
\caption{SUPERB downstream results on content, speaker,
paralinguistic, and semantic tasks. Params denotes the encoder parameter
count. \textbf{Bold}: best; \underline{underlined}: second-best.}
\label{tab:superb-main-results}
\resizebox{\textwidth}{!}{%
\begin{tabular}{llrrrrrrrrrrrr}
\toprule
 & & \multicolumn{4}{c}{Content} & \multicolumn{3}{c}{Speaker} & \multicolumn{1}{c}{Paraling.} & \multicolumn{4}{c}{Semantics} \\
\cmidrule(lr){3-6} \cmidrule(lr){7-9} \cmidrule(lr){10-10} \cmidrule(lr){11-14}
\multicolumn{1}{c}{Model} & \multicolumn{1}{c}{Params} & PR & ASR & KS & QbE & SID & ASV & SD & ER & IC & \multicolumn{2}{c}{SF} & ST \\
\cmidrule(lr){3-3} \cmidrule(lr){4-4} \cmidrule(lr){5-5} \cmidrule(lr){6-6}
\cmidrule(lr){7-7} \cmidrule(lr){8-8} \cmidrule(lr){9-9}
\cmidrule(lr){10-10}
\cmidrule(lr){11-11} \cmidrule(lr){12-13} \cmidrule(lr){14-14}
 & & PER$\downarrow$ & WER$\downarrow$ & Acc.$\uparrow$ & MTWV$\uparrow$ & Acc.$\uparrow$ & EER$\downarrow$ & DER$\downarrow$ & Acc.$\uparrow$ & Acc.$\uparrow$ & F1$\uparrow$ & CER$\downarrow$ & BLEU$\uparrow$ \\
\midrule
FBANK & -- & 82.1 & 23.1 & 8.3 & 0.0043 & 0.1 & 10.7 & 11.3 & 28.4 & 4.6 & 63.0 & 59.5 & 2.6 \\
wav2vec 2.0 Base & 95.04M & 6.4 & 6.5 & 96.0 & 0.0634 & 66.5 & 5.8 & 6.6 & 62.5 & 92.6 & 87.0 & 26.5 & 14.9 \\
HuBERT Base & 94.70M & 5.8 & 6.4 & 96.4 & \underline{0.0748} & \underline{67.3} & 5.6 & 6.2 & 65.1 & 98.2 & 88.0 & 25.1 & 15.8 \\
WavLM Base & 94.38M & 5.6 & 6.2 & \underline{97.0} & \textbf{0.0813} & 62.3 & 5.8 & 5.4 & 64.9 & 98.5 & \underline{89.0} & \underline{24.2} & 16.3 \\
data2vec Base & 93.16M & \textbf{3.8} & 5.0 & 95.9 & 0.0609 & 54.9 & 7.0 & 7.0 & 65.4 & 98.6 & \underline{89.0} & \underline{23.4} & \underline{17.5} \\
data2vec 2.0 Base & 93.16M & \underline{4.6} & \textbf{4.8} & 96.8 & 0.0556 & 45.5 & \underline{5.5} & 6.3 & 63.9 & \underline{99.0} & \underline{89.0} & \underline{23.4} & \textbf{17.8} \\
\midrule
\rowcolor{olivehighlight}
OLIVE-A (Mix) & 93.16M & 6.4 & 6.4 & \underline{97.0} & 0.0390 & 66.9 & 6.0 & \underline{4.6} & \underline{67.1} & \textbf{99.1} & 88.0 & 25.2 & 16.5 \\
\rowcolor{olivehighlight}
OLIVE-A (Mix+Gain) & 93.16M & 5.2 & 6.1 & \textbf{97.3} & 0.0459 & 63.0 & \textbf{5.5} & \textbf{4.3} & \textbf{67.8} & 98.9 & 88.0 & 24.4 & 16.8 \\
\rowcolor{olivehighlight}
OLIVE-J (Mask-only) & 93.16M & 4.7 & \underline{4.9} & \underline{97.0} & 0.0480 & 64.6 & 5.7 & 6.8 & 66.9 & 98.6 & \textbf{89.4} & \textbf{22.3} & 17.3 \\
\rowcolor{olivehighlight}
OLIVE-J (Mix) & 93.16M & 6.3 & 6.4 & \textbf{97.3} & 0.0445 & \textbf{83.1} & 5.8 & \underline{4.6} & 64.6 & 98.1 & 88.0 & 25.8 & 15.4 \\
\bottomrule
\end{tabular}%
}
\end{table*}

\begin{table*}[t]
\centering
\footnotesize
\setlength{\tabcolsep}{4pt}
\caption{SUPERB downstream results on generation tasks, with overall
SUPERB$_s$ and WavLM$_s$ scores. \textbf{Bold}: best; \underline{underlined}: second-best.}
\label{tab:superb-main-results-gen}
\resizebox{.8\textwidth}{!}{%
\begin{tabular}{lrrrrrrrr}
\toprule
\multicolumn{1}{c}{Model} & \multicolumn{2}{c}{SE} & SS & \multicolumn{3}{c}{VC} & \multicolumn{2}{c}{Overall} \\
\cmidrule(lr){2-3} \cmidrule(lr){4-4} \cmidrule(lr){5-7} \cmidrule(lr){8-9}
 & PESQ$\uparrow$ & STOI$\uparrow$ & SI-SDRi$\uparrow$ & MCD$\downarrow$ & WER$\downarrow$ & ASV$\uparrow$ & SUPERB$_s$$\uparrow$ & WavLM$_s$$\uparrow$ \\
\midrule
FBANK & 2.83 & 94.3 & 9.2 & 8.21 & 41.40 & 87.0 & 0 & 34.2 \\
data2vec-SG Base$^\dagger$ & 2.59 & 94.0 & 10.8 & -- & -- & -- & -- & -- \\
wav2vec 2.0 Base & 2.92 & 94.8 & 10.3 & 7.50 & 11.30 & 97.8 & 795 & 65.5 \\
HuBERT Base & 3.00 & 94.9 & 9.9 & 7.48 & 10.90 & 98.0 & 836 & 66.5 \\
WavLM Base & 2.99 & 94.9 & 10.8 & 7.44 & \textbf{8.95} & 98.2 & 875 & 66.6 \\
data2vec Base & 2.94 & 94.8 & 9.9 & \textbf{7.08} & \underline{9.65} & \textbf{99.5} & 805 & 65.9 \\
data2vec 2.0 Base & 2.96 & 94.9 & 10.4 & \underline{7.28} & 10.40 & \textbf{99.5} & 835 & 65.3 \\
\midrule
\rowcolor{olivehighlight}
OLIVE-A (Mix) & 3.05 & \textbf{95.2} & \underline{12.2} & 7.71 & 12.20 & 95.5 & 884 & 66.7 \\
\rowcolor{olivehighlight}
OLIVE-A (Mix+Gain) & \underline{3.06} & \textbf{95.2} & \underline{12.2} & 7.59 & 12.10 & 96.2 & \textbf{909} & \underline{66.8} \\
\rowcolor{olivehighlight}
OLIVE-J (Mask-only) & 2.94 & 94.8 & 10.5 & \underline{7.28} & 11.10 & \underline{99.0} & 835 & \underline{66.8} \\
\rowcolor{olivehighlight}
OLIVE-J (Mix) & \textbf{3.10} & \underline{95.1} & \textbf{12.3} & 7.61 & 11.60 & 98.2 & \underline{908} & \textbf{67.6} \\
\bottomrule
\end{tabular}%
}
\end{table*}

%% file: tables/hifigan_compact.tex
\begin{table*}[t]
\centering
\setlength{\tabcolsep}{3.5pt}
\scriptsize
\caption{Waveform reconstruction on LibriSpeech test-clean with matched and integrated HiFi-GAN V2 vocoders (means; 95\% CIs in Appendix~\ref{app:reconstruction_ci}). \textbf{Bold}: best; \underline{underlined}: second-best.}
\label{tab:resynthesis_main}
\begin{tabular}{lcccccccccc}
\toprule
Model & Mel-L1 $\downarrow$ & MCD $\downarrow$ & LSD $\downarrow$ & $F_0$ MAE $\downarrow$ & STOI $\uparrow$ & PESQ $\uparrow$ & ViSQOL $\uparrow$ & SI-SDR $\uparrow$ & SNR $\uparrow$ & UTMOS $\uparrow$ \\
\midrule
Reference audio & & & & & & & & & & 4.09 \\
\midrule
Mel spectrogram & 0.677 & 6.24 & 0.829 & 14.4 & 0.833 & 1.42 & 3.59 & $-18.02$ & $-2.60$ & 2.98 \\
wav2vec 2.0 Base & 0.474 & 5.18 & 0.604 & 11.6 & 0.912 & 2.19 & 4.04 & $-7.53$ & $-0.42$ & 3.58 \\
HuBERT Base & 0.475 & 5.05 & 0.604 & 13.3 & 0.914 & 2.11 & 4.05 & $-11.78$ & $-1.57$ & 3.63 \\
WavLM Base & 0.470 & 5.01 & 0.599 & 13.1 & 0.917 & 2.21 & 4.05 & $-10.93$ & $-1.31$ & 3.68 \\
data2vec Base & 0.467 & 4.90 & 0.594 & 12.8 & 0.902 & 2.10 & 4.06 & $-10.64$ & $-1.15$ & 3.50 \\
data2vec 2.0 Base & 0.434 & 4.78 & 0.558 & 13.0 & 0.917 & 2.21 & 4.10 & $-9.67$ & $-1.11$ & 3.63 \\
\midrule
\rowcolor{olivehighlight}
\multicolumn{11}{l}{\emph{Frozen feature-conditioned vocoder}} \\
\rowcolor{olivehighlight}
OLIVE-A (Mix+Gain) & 0.446 & 4.69 & 0.566 & 11.6 & 0.929 & 2.62 & 4.18 & $-3.36$ & 1.06 & 3.70 \\
\rowcolor{olivehighlight}
OLIVE-J (Mask-only) & \textbf{0.410} & \underline{4.43} & \textbf{0.525} & 10.8 & \underline{0.935} & 2.81 & \underline{4.26} & 0.97 & 3.31 & \textbf{3.86} \\
\rowcolor{olivehighlight}
OLIVE-J (Mix) & \underline{0.421} & \textbf{4.35} & \underline{0.528} & \textbf{9.6} & \textbf{0.942} & \textbf{3.06} & \textbf{4.34} & \textbf{2.94} & \textbf{4.69} & \underline{3.83} \\
\midrule
\rowcolor{oliveintegrated}
\multicolumn{11}{l}{\emph{Integrated vocoder}} \\
\rowcolor{oliveintegrated}
OLIVE-J (Mask-only) & 0.476 & 4.68 & 0.593 & 11.1 & 0.911 & 2.56 & 4.14 & 0.72 & 3.08 & 3.74 \\
\rowcolor{oliveintegrated}
OLIVE-J (Mix) & 0.579 & 4.56 & 0.688 & \underline{10.0} & 0.921 & \underline{2.88} & 4.21 & \underline{2.50} & \underline{4.25} & 3.78 \\
\bottomrule
\end{tabular}
\end{table*}

%% file: figures/knob_pareto.tex
\definecolor{knobD2v2}{HTML}{D08A2E}
\definecolor{knobAMix}{HTML}{4B9B68}
\definecolor{knobAGain}{HTML}{1F7F5C}
\definecolor{knobJMask}{HTML}{A44F89}
\definecolor{knobJMix}{HTML}{C84A3A}
\tikzset{
    augarrow/.style={-{Stealth[length=5pt]}, line width=0.9pt, black!60, dashed, shorten >=3.2pt, shorten <=0pt},
    recarrow/.style={-{Stealth[length=5pt]}, line width=0.9pt, black!80, shorten >=3.2pt, shorten <=0pt},
    frontier/.style={black!12, line width=5pt, line cap=round, line join=round},
}
\pgfplotsset{
    knobaxis/.style={
        width=7.4cm, height=4.3cm,
        xlabel={ASR WER test-other (\%) $\downarrow$},
        xmin=10.5, xmax=16.5,
        tick label style={font=\scriptsize},
        label style={font=\footnotesize},
        title style={font=\footnotesize\bfseries},
        grid=major, grid style={black!8},
        axis line style={black!40},
        clip=false,
    },
}
\begin{tikzpicture}[font=\sffamily]
\begin{axis}[knobaxis, name=left,
    title={Downstream performance},
    ylabel={Speaker + Generation\\score $\uparrow$}, ylabel style={align=center},
    ymin=580, ymax=1015, ytick={600,700,800,900}]
    \addplot[frontier] coordinates {(11.1,691.4) (11.4,698.0) (14.4,869.2) (15.7,937.6)};
    \addplot[only marks, mark=*, mark size=1.6pt, black!35] coordinates {(16.0,669.9) (15.6,702.5) (15.0,756.6) (12.3,614.4)};
    \node[font=\tiny, black!55, anchor=north] at (axis cs:16.0,669.9) {w2v2};
    \node[font=\tiny, black!55, anchor=south] at (axis cs:15.6,704) {HuBERT};
    \node[font=\tiny, black!55, anchor=south] at (axis cs:15.0,758) {WavLM};
    \node[font=\tiny, black!55, anchor=west] at (axis cs:12.4,614.4) {data2vec};
    \draw[augarrow] (axis cs:11.1,691.4) to[bend right=18] (axis cs:14.4,869.2);
    \draw[augarrow] (axis cs:11.4,698.0) to[bend left=18] (axis cs:15.7,937.6);
    \draw[recarrow, -{Stealth[length=3.5pt]}, shorten >=2.6pt, shorten <=2pt] (axis cs:11.1,691.4) -- (axis cs:11.4,698.0);
    \draw[recarrow] (axis cs:14.4,869.2) -- (axis cs:15.7,937.6);
    \addplot[only marks, mark=square*, mark size=2.6pt, knobD2v2] coordinates {(11.1,691.4)};
    \addplot[only marks, mark=*, mark size=2.8pt, knobAMix] coordinates {(14.4,869.2)};
    \addplot[only marks, mark=triangle*, mark size=3.2pt, knobJMask] coordinates {(11.4,698.0)};
    \addplot[only marks, mark=pentagon*, mark size=3.2pt, knobJMix] coordinates {(15.7,937.6)};
    \node[font=\tiny, knobD2v2!80!black, anchor=north] at (axis cs:11.1,684) {data2vec 2.0};
    \node[font=\tiny, knobJMask!80!black, anchor=north west] at (axis cs:11.75,692) {J (Mask-only)};
    \node[font=\tiny, knobAMix!80!black, anchor=west] at (axis cs:14.55,866) {A (Mix)};
    \node[font=\tiny, knobJMix!80!black, anchor=south] at (axis cs:15.7,944) {J (Mix)};
\end{axis}
\begin{axis}[knobaxis, at={(left.east)}, anchor=west, xshift=1.3cm,
    title={Waveform reconstruction},
    ylabel={PESQ $\uparrow$},
    ymin=2.0, ymax=3.28]
    \addplot[frontier] coordinates {(11.1,2.21) (11.4,2.81) (15.7,3.06)};
    \addplot[only marks, mark=*, mark size=1.6pt, black!35] coordinates {(16.0,2.19) (15.6,2.11) (15.0,2.21) (12.3,2.10)};
    \node[font=\tiny, black!55, anchor=south] at (axis cs:16.0,2.19) {w2v2};
    \node[font=\tiny, black!55, anchor=east] at (axis cs:15.5,2.11) {HuBERT};
    \node[font=\tiny, black!55, anchor=south] at (axis cs:15.0,2.21) {WavLM};
    \node[font=\tiny, black!55, anchor=west] at (axis cs:12.4,2.10) {data2vec};
    \draw[augarrow] (axis cs:11.1,2.21) -- (axis cs:14.0,2.62);
    \draw[augarrow] (axis cs:11.4,2.81) -- (axis cs:15.7,3.06);
    \draw[recarrow] (axis cs:11.1,2.21) -- (axis cs:11.4,2.81);
    \draw[recarrow] (axis cs:14.0,2.62) -- (axis cs:15.7,3.06);
    \addplot[only marks, mark=square*, mark size=2.6pt, knobD2v2] coordinates {(11.1,2.21)};
    \addplot[only marks, mark=diamond*, mark size=3.2pt, knobAGain] coordinates {(14.0,2.62)};
    \addplot[only marks, mark=triangle*, mark size=3.2pt, knobJMask] coordinates {(11.4,2.81)};
    \addplot[only marks, mark=pentagon*, mark size=3.2pt, knobJMix] coordinates {(15.7,3.06)};
    \node[font=\tiny, knobD2v2!80!black, anchor=north] at (axis cs:11.1,2.19) {data2vec 2.0};
    \node[font=\tiny, knobJMask!80!black, anchor=south west] at (axis cs:11.45,2.83) {J (Mask-only)};
    \node[font=\tiny, knobAGain!80!black, anchor=north west] at (axis cs:14.05,2.60) {A (Mix+Gain)};
    \node[font=\tiny, knobJMix!80!black, anchor=south] at (axis cs:15.7,3.09) {J (Mix)};
\end{axis}
\node[font=\scriptsize, anchor=north] at ($(left.south east)!0.5!(left.south east)+(0.65cm,-0.75cm)$) {
    \tikz[baseline=-0.55ex]{\draw[recarrow, shorten >=0pt, shorten <=0pt] (0,0) -- (0.6,0);}\;+ reconstruction
    \qquad
    \tikz[baseline=-0.55ex]{\draw[augarrow, shorten >=0pt, shorten <=0pt] (0,0) -- (0.6,0);}\;+ waveform view augmentation
    \qquad
    \tikz[baseline=-0.55ex]{\draw[frontier] (0,0) -- (0.5,0);}\;Pareto front
};
\end{tikzpicture}

%% file: figures/superb_radar_macros.tex
\definecolor{taskWav2vec}{HTML}{7F8C8D}
\definecolor{taskHuBERT}{HTML}{8E6AB8}
\definecolor{taskWavLM}{HTML}{3B6FB6}
\definecolor{taskData2vec}{HTML}{D08A2E}
\definecolor{taskOliveMix}{HTML}{4B9B68}
\definecolor{taskOliveGain}{HTML}{1F7F5C}
\definecolor{taskOliveJ}{HTML}{C84A3A}
\definecolor{taskOliveMask}{HTML}{A44F89}

\providecommand{\TaskR}{2.24}
\providecommand{\TaskRadarPlot}[2]{%
    \path[fill=#1, fill opacity=0.08] #2 -- cycle;
    \draw[#1, line width=0.75pt] #2 -- cycle;
}
\providecommand{\TaskAxes}[1]{%
    \node[font=\small\sffamily\bfseries, align=center] at (0,3.18) {#1};
    \foreach \s in {0.25,0.50,0.75,1.00} {
        \draw[black!12, line width=0.4pt]
            (90:{\TaskR*\s}) --
            (64.286:{\TaskR*\s}) --
            (38.571:{\TaskR*\s}) --
            (12.857:{\TaskR*\s}) --
            (-12.857:{\TaskR*\s}) --
            (-38.571:{\TaskR*\s}) --
            (-64.286:{\TaskR*\s}) --
            (-90:{\TaskR*\s}) --
            (-115.714:{\TaskR*\s}) --
            (-141.429:{\TaskR*\s}) --
            (-167.143:{\TaskR*\s}) --
            (-192.857:{\TaskR*\s}) --
            (-218.571:{\TaskR*\s}) --
            (-244.286:{\TaskR*\s}) -- cycle;
    }
    \foreach \a in {90,64.286,38.571,12.857,-12.857,-38.571,-64.286,-90,-115.714,-141.429,-167.143,-192.857,-218.571,-244.286} {
        \draw[black!16, line width=0.35pt] (0,0) -- (\a:\TaskR);
    }
    \node[font=\small\sffamily, align=center, text=blue!55!black] at (90:2.66) {PR};
    \node[font=\small\sffamily, align=center, text=blue!55!black] at (64.286:2.66) {ASR};
    \node[font=\small\sffamily, align=center, text=blue!55!black] at (38.571:2.66) {KS};
    \node[font=\small\sffamily, align=center, text=blue!55!black] at (12.857:2.66) {QbE};
    \node[font=\small\sffamily, align=center, text=violet!65!black] at (-12.857:2.74) {SID};
    \node[font=\small\sffamily, align=center, text=violet!65!black] at (-38.571:2.74) {ASV};
    \node[font=\small\sffamily, align=center, text=violet!65!black] at (-64.286:2.66) {SD};
    \node[font=\small\sffamily, align=center, text=orange!75!black] at (-90:2.58) {ER};
    \node[font=\small\sffamily, align=center, text=teal!60!black] at (-115.714:2.72) {IC};
    \node[font=\small\sffamily, align=center, text=teal!60!black] at (-141.429:2.78) {SF};
    \node[font=\small\sffamily, align=center, text=teal!60!black] at (-167.143:2.68) {ST};
    \node[font=\small\sffamily, align=center, text=red!60!black] at (-192.857:2.66) {SE};
    \node[font=\small\sffamily, align=center, text=red!60!black] at (-218.571:2.70) {SS};
    \node[font=\small\sffamily, align=center, text=red!60!black] at (-244.286:2.66) {VC};
    \node[font=\small\sffamily, text=black!55, fill=white, inner sep=1pt] at (92:{\TaskR*0.50}) {500};
    \node[font=\small\sffamily, text=black!55, fill=white, inner sep=1pt] at (92:{\TaskR*1.00}) {1000};
}

\providecommand{\CatR}{2.24}
\providecommand{\CatRadarPlot}[6]{%
    \path[fill=#1, fill opacity=0.10]
        (90:{\CatR*#2/1000}) --
        (18:{\CatR*#3/1000}) --
        (-54:{\CatR*#4/1000}) --
        (-126:{\CatR*#5/1000}) --
        (162:{\CatR*#6/1000}) -- cycle;
    \draw[#1, line width=0.85pt]
        (90:{\CatR*#2/1000}) --
        (18:{\CatR*#3/1000}) --
        (-54:{\CatR*#4/1000}) --
        (-126:{\CatR*#5/1000}) --
        (162:{\CatR*#6/1000}) -- cycle;
}
\providecommand{\CatAxes}[1]{%
    \node[font=\normalsize\sffamily\bfseries, align=center] at (0,3.18) {#1};
    \foreach \s in {0.25,0.50,0.75,1.00} {
        \draw[black!13, line width=0.45pt]
            (90:{\CatR*\s}) --
            (18:{\CatR*\s}) --
            (-54:{\CatR*\s}) --
            (-126:{\CatR*\s}) --
            (162:{\CatR*\s}) -- cycle;
    }
    \foreach \a in {90,18,-54,-126,162} {
        \draw[black!16, line width=0.45pt] (0,0) -- (\a:\CatR);
    }
    \node[font=\small\sffamily, align=center] at (90:2.64) {Content};
    \node[font=\small\sffamily, align=center] at (18:2.76) {Speaker};
    \node[font=\small\sffamily, align=center] at (-54:2.80) {Paralinguistics};
    \node[font=\small\sffamily, align=center] at (-126:2.82) {Semantics};
    \node[font=\small\sffamily, align=center] at (162:2.78) {Generation};
    \node[font=\small\sffamily, text=black!55, fill=white, inner sep=1pt] at (92:{\CatR*0.50}) {500};
    \node[font=\small\sffamily, text=black!55, fill=white, inner sep=1pt] at (92:{\CatR*1.00}) {1000};
}

\providecommand{\LegendItem}[2]{%
    \tikz[baseline=-0.55ex]{\draw[#1, line width=1.8pt] (0,0) -- (0.42,0);}\,#2%
}

%% file: figures/superb_category_2x2.tex
\input{figures/superb_radar_macros}
\begin{tikzpicture}[font=\sffamily]
    \draw[black!13, line width=0.45pt] (90:0.448) -- (18:0.448) -- (-54:0.448) -- (-126:0.448) -- (162:0.448) -- cycle;
    \draw[black!13, line width=0.45pt] (90:0.896) -- (18:0.896) -- (-54:0.896) -- (-126:0.896) -- (162:0.896) -- cycle;
    \draw[black!13, line width=0.45pt] (90:1.344) -- (18:1.344) -- (-54:1.344) -- (-126:1.344) -- (162:1.344) -- cycle;
    \draw[black!13, line width=0.45pt] (90:1.792) -- (18:1.792) -- (-54:1.792) -- (-126:1.792) -- (162:1.792) -- cycle;
    \draw[black!13, line width=0.45pt] (90:2.240) -- (18:2.240) -- (-54:2.240) -- (-126:2.240) -- (162:2.240) -- cycle;
    \draw[black!16, line width=0.45pt] (0,0) -- (90:2.24);
    \draw[black!16, line width=0.45pt] (0,0) -- (18:2.24);
    \draw[black!16, line width=0.45pt] (0,0) -- (-54:2.24);
    \draw[black!16, line width=0.45pt] (0,0) -- (-126:2.24);
    \draw[black!16, line width=0.45pt] (0,0) -- (162:2.24);
    \path[fill=taskData2vec, fill opacity=0.08] (90:1.848) -- (18:1.137) -- (-54:1.797) -- (-126:2.205) -- (162:0.578) -- cycle;
    \draw[taskData2vec, line width=1.0pt] (90:1.848) -- (18:1.137) -- (-54:1.797) -- (-126:2.205) -- (162:0.578) -- cycle;
    \path[fill=taskOliveMix, fill opacity=0.08] (90:1.486) -- (18:1.741) -- (-54:2.160) -- (-126:2.014) -- (162:1.567) -- cycle;
    \draw[taskOliveMix, line width=1.0pt] (90:1.486) -- (18:1.741) -- (-54:2.160) -- (-126:2.014) -- (162:1.567) -- cycle;
    \path[fill=taskOliveMask, fill opacity=0.08] (90:1.733) -- (18:1.316) -- (-54:2.138) -- (-126:2.183) -- (162:0.458) -- cycle;
    \draw[taskOliveMask, line width=1.0pt] (90:1.733) -- (18:1.316) -- (-54:2.138) -- (-126:2.183) -- (162:0.458) -- cycle;
    \path[fill=taskOliveJ, fill opacity=0.08] (90:1.571) -- (18:2.090) -- (-54:1.876) -- (-126:1.879) -- (162:1.831) -- cycle;
    \draw[taskOliveJ, line width=1.0pt] (90:1.571) -- (18:2.090) -- (-54:1.876) -- (-126:1.879) -- (162:1.831) -- cycle;
    \node[font=\small\sffamily] at (90:2.62) {Content};
    \node[font=\small\sffamily] at (18:2.76) {Speaker};
    \node[font=\small\sffamily] at (-54:2.8) {Paralinguistics};
    \node[font=\small\sffamily] at (-126:2.8) {Semantics};
    \node[font=\small\sffamily] at (162:2.78) {Generation};
    \node[font=\scriptsize\sffamily, text=black!55, fill=white, inner sep=1pt] at (92:0.448) {600};
    \node[font=\scriptsize\sffamily, text=black!55, fill=white, inner sep=1pt] at (92:1.344) {800};
    \node[font=\scriptsize\sffamily, text=black!55, fill=white, inner sep=1pt] at (92:2.240) {1000};
\end{tikzpicture}

%% file: tables/model_pretraining_details.tex
\begin{table*}[!ht]
\centering
\caption{OLIVE model architecture and pre-training configuration.}
\label{tab:model_pretraining_details}
\resizebox{\textwidth}{!}{
\begin{tabular}{ll}
\toprule
\textbf{Component} & \textbf{Setting} \\
\midrule
\multicolumn{2}{l}{\textbf{Optimization and Pre-training Setup}} \\
Implementation & fairseq \\
Pre-training data & LibriSpeech 960h unlabeled training set \\
Sampling rate & 16~kHz \\
Model size & Base; 12 Transformer layers, hidden size $d=768$ \\
Small ablations & Hidden size $d=384$; 100k updates \\
Optimizer & Adam, $\beta_1=0.9$, $\beta_2=0.98$, $\epsilon=10^{-6}$ \\
Weight decay & 0.01 \\
Peak learning rate & $7.5\times 10^{-4}$ \\
Learning-rate schedule & cosine decay after 8k warmup updates \\
Training updates & 400k \\
Batching & 1M raw samples/GPU; gradient accumulation 8 \\
Hardware & 2 NVIDIA H100 GPUs \\
Average realized batch size & 72.7 utterances/update \\
Gradient clipping & disabled \\
Precision & FP16 analysis branch; FP32 synthesis branch \\
\midrule
\multicolumn{2}{l}{\textbf{Encoder Architecture}} \\
Waveform encoder & $[(512,10,5)] + 4 \times [(512,3,2)] + 2 \times [(512,2,2)]$ \\
Waveform stride & 320 samples / 20~ms \\
Effective receptive field & 400 samples / 25~ms \\
Projection & 512 $\rightarrow$ 768 \\
Positional encoder & 5-layer convolutional positional encoder (kernel size 19 per layer; effective width 95; 16 groups) \\
Transformer encoder & 12 layers, 12 heads, hidden size $d=768$, MLP ratio 4 \\
Dropout & encoder 0.1; attention 0.1; post-MLP 0.1; activation 0.0; input 0.0 \\
Layer drop & 0.05 \\
Teacher target layers & top $K=8$ layers after instance normalization \\
Prediction head & 4-layer 1D convolutional decoder; dim 384; groups 16; kernel size 7; dropout 0.1 \\
Masking & inverse-block masking; mask ratio 0.5; block width 5; adjustment 0.05; noise std.\ 0.01 \\
Masked views per sample & 8 \\
EMA schedule & 0.999 $\rightarrow$ 0.99999 over 75k updates \\
\midrule
\multicolumn{2}{l}{\textbf{OLIVE-J Synthesis Branch}} \\
Vocoder & HiFi-GAN V2 generator \\
Conditioning representation & local encoder output \\
Conditioning dimension & 768 \\
Internal bottleneck & 128 \\
Upsampling rates / kernels & $[10,8,2,2]$ / $[20,16,4,4]$ \\
Initial upsampling width & 128 \\
Residual block kernels / dilations & $[3,7,11]$ / $(1,3,5)$ \\
Discriminators & HiFi-GAN MPD + MSD \\
MPD periods & $\{2,3,5,7,11\}$ \\
MSD branches & 3 sub-discriminators; spectral normalization in first branch \\
Synthesis segment length & 7040 samples \\
Mel analysis & 80 bins; FFT 1024; window 1024; hop 320 \\
Loss weights & $\lambda_{\mathrm{synthesis}}=25$, $\lambda_{\mathrm{fm}}=1$, $\lambda_{\mathrm{mel}}=45$ \\
\bottomrule
\end{tabular}
}
\end{table*}

%% file: tables/parameter_breakdown.tex
\begin{table*}[!ht]
\centering
\caption{Parameter breakdown for OLIVE-J. The HiFi-GAN discriminators are used
only during pre-training and are discarded afterwards.}
\label{tab:parameter_breakdown}
\resizebox{.5\textwidth}{!}{
\begin{tabular}{lr}
\toprule
Component & Parameters \\
\midrule
\multicolumn{2}{l}{\textbf{Shared local encoder}} \\
Convolutional waveform encoder & 4.2M \\
Feature projection (512 $\rightarrow$ 768) & 0.4M \\
\midrule
\multicolumn{2}{l}{\textbf{Analysis branch}} \\
Relative positional convolutional encoder & 3.5M \\
Transformer encoder & 85.1M \\
Prediction decoder & 0.6M \\
\midrule
\multicolumn{2}{l}{\textbf{Synthesis branch}} \\
HiFi-GAN generator & 1.1M \\
HiFi-GAN multi-period discriminator & 41.1M \\
HiFi-GAN multi-scale discriminator & 29.6M \\
\midrule
\multicolumn{2}{l}{\textbf{Totals (training-time)}} \\
Shared local encoder & 4.6M \\
Analysis branch & 89.2M \\
Synthesis branch & 71.8M \\
Full joint model & 165.6M \\
\bottomrule
\end{tabular}
}
\end{table*}

%% file: tables/d2vsg_relative.tex
\begin{table}[!ht]
\centering
\caption{Relative comparison on SUPERB generation tasks. $\Delta$ is the difference
from data2vec Base within the same evaluation.}
\label{tab:d2vsg_relative}
\small
\begin{tabular}{lrrr}
\toprule
 & \multicolumn{2}{c}{SE} & SS \\
\cmidrule(lr){2-3} \cmidrule(lr){4-4}
Model & PESQ$\uparrow$ & STOI$\uparrow$ & SI-SDRi$\uparrow$ \\
\midrule
\multicolumn{4}{l}{\textit{Reported by \citet{wang2023data2vec}}} \\
FBANK & 2.55 & 93.6 & 9.2 \\
data2vec Base & 2.57 & 93.9 & 9.8 \\
data2vec-SG Base & 2.59 & 94.0 & 10.8 \\
\quad $\Delta$ vs.\ data2vec & $+0.02$ & $+0.1$ & $+1.0$ \\
\midrule
\multicolumn{4}{l}{\textit{Our evaluation}} \\
FBANK & 2.83 & 94.3 & 9.2 \\
data2vec Base & 2.94 & 94.8 & 9.9 \\
OLIVE-J (Mask-only) & 2.94 & 94.8 & 10.5 \\
\quad $\Delta$ vs.\ data2vec & $0.00$ & $0.0$ & $+0.6$ \\
OLIVE-J (Mix) & 3.10 & 95.1 & 12.3 \\
\quad $\Delta$ vs.\ data2vec & $+0.16$ & $+0.3$ & $+2.4$ \\
\bottomrule
\end{tabular}
\end{table}

%% file: tables/loss_weight_ablation.tex
\begin{table*}[!ht]
\centering
\caption{Synthesis loss weight ablation on a representative subset of SUPERB downstream tasks. All models are 100k-step OLIVE-J (Mix) variants trained on LibriSpeech 960h. \textbf{Bold}: best; \underline{underlined}: second-best.}
\label{tab:loss_weight_subset}
\resizebox{\textwidth}{!}{
\begin{tabular}{lccccccc}
\toprule
$\lambda_{\mathrm{synthesis}}$ & ASV EER$\downarrow$ & ER Acc$\uparrow$ & PR PER$\downarrow$ & QbE $\mathrm{maxTWV}\uparrow$ & VC MCD$\downarrow$ & VC WER$\downarrow$ & VC $F_0$ Corr$\uparrow$ \\
\midrule
$1$  & \textbf{0.0536} & 0.666 & \textbf{0.0647} & 0.0515 & \underline{7.55} & \underline{10.6} & 0.282 \\
$25$ & \underline{0.0581} & \textbf{0.673} & \underline{0.0652} & \underline{0.0535} & \textbf{7.52} & \textbf{10.2} & \underline{0.284} \\
$50$ & 0.0643 & \underline{0.667} & 0.0667 & \textbf{0.0551} & \underline{7.55} & 12.1 & \textbf{0.289} \\
\bottomrule
\end{tabular}
}
\end{table*}

\begin{table}[!ht]
\centering
\caption{Synthesis loss weight ablation on waveform reconstruction quality using the integrated OLIVE-J (Mix) vocoder, evaluated on the LibriSpeech dev-clean set. All models are 100k-step OLIVE-J (Mix) variants. \textbf{Bold}: best; \underline{underlined}: second-best.}
\label{tab:loss_weight_reconstruction}
\begin{tabular}{lcccccc}
\toprule
$\lambda_{\mathrm{synthesis}}$ & Mel-L1$\downarrow$ & MCD$\downarrow$ & STOI$\uparrow$ & PESQ$\uparrow$ & $F_0$ MAE$\downarrow$ & UTMOS$\uparrow$  \\
\midrule
$1$  & 1.149 & 5.06 & 0.873 & 2.27 & 6.74 & 3.51  \\
$25$ & 1.086 & \textbf{4.72} & 0.891 & 2.66 & 5.01 & \textbf{3.88}  \\
$50$ & \textbf{1.007} & 4.93 & \textbf{0.894} & \textbf{2.79} & \textbf{4.67} & 3.78 \\
\bottomrule
\end{tabular}
\end{table}

%% file: figures/loss_weight_pareto.tex
\definecolor{lwPoint}{HTML}{C84A3A}
\pgfplotsset{
    lwaxis/.style={
        width=6.2cm, height=4.6cm,
        xlabel={PR PER (\%) $\downarrow$},
        xmin=6.42, xmax=6.72,
        xtick={6.45,6.50,6.55,6.60,6.65,6.70},
        x tick label style={/pgf/number format/.cd, fixed, fixed zerofill, precision=2},
        tick label style={font=\scriptsize},
        label style={font=\footnotesize},
        grid=major, grid style={black!8},
        axis line style={black!40},
        clip=false,
    },
}
\begin{tikzpicture}[font=\sffamily]
\begin{axis}[lwaxis, name=left, ylabel={PESQ $\uparrow$}, ymin=2.2, ymax=2.85]
    \addplot[black!45, line width=0.8pt, -{Stealth[length=4pt]}, mark=none] coordinates {(6.47,2.27) (6.52,2.66)};
    \addplot[black!45, line width=0.8pt, -{Stealth[length=4pt]}, mark=none] coordinates {(6.52,2.66) (6.67,2.79)};
    \addplot[only marks, mark=*, mark size=2.4pt, lwPoint] coordinates {(6.47,2.27) (6.52,2.66) (6.67,2.79)};
    \node[font=\scriptsize, anchor=west] at (axis cs:6.48,2.27) {$\lambda=1$};
    \node[font=\scriptsize, anchor=west] at (axis cs:6.53,2.66) {$\lambda=25$};
    \node[font=\scriptsize, anchor=north] at (axis cs:6.67,2.77) {$\lambda=50$};
\end{axis}
\begin{axis}[lwaxis, at={(left.east)}, anchor=west, xshift=2.0cm, ylabel={MCD $\downarrow$}, ymin=4.65, ymax=5.12]
    \addplot[black!45, line width=0.8pt, -{Stealth[length=4pt]}, mark=none] coordinates {(6.47,5.06) (6.52,4.72)};
    \addplot[black!45, line width=0.8pt, -{Stealth[length=4pt]}, mark=none] coordinates {(6.52,4.72) (6.67,4.93)};
    \addplot[only marks, mark=*, mark size=2.4pt, lwPoint] coordinates {(6.47,5.06) (6.52,4.72) (6.67,4.93)};
    \node[font=\scriptsize, anchor=west] at (axis cs:6.48,5.06) {$\lambda=1$};
    \node[font=\scriptsize, anchor=west] at (axis cs:6.53,4.72) {$\lambda=25$};
    \node[font=\scriptsize, anchor=north] at (axis cs:6.67,4.915) {$\lambda=50$};
\end{axis}
\end{tikzpicture}

%% file: tables/variant_selection_ablation.tex
\begin{table*}[!ht]
\centering
\caption{Small-model variant ablation used to select which analysis-only
OLIVE augmentations to train at full scale. All models use hidden size $d=384$ and
are trained for 100k updates. $\mathrm{Avg}$ is the mean of all metrics after
normalizing each to $[0,100]$ across OLIVE variants, with lower-is-better
metrics inverted. \textbf{Bold}: best; \underline{underlined}: second-best.}
\label{tab:variant_selection_ablation}
\resizebox{\textwidth}{!}{
\begin{tabular}{lccccccccccc}
\toprule
Model & PR & ASV & SID & ER & \multicolumn{3}{c}{VC} & \multicolumn{3}{c}{SE} & Avg \\
\cmidrule(lr){2-2}\cmidrule(lr){3-3}\cmidrule(lr){4-4}\cmidrule(lr){5-5}\cmidrule(lr){6-8}\cmidrule(lr){9-11}\cmidrule(lr){12-12}
& PER$\downarrow$ & EER$\downarrow$ & Acc$\uparrow$ & Acc$\uparrow$ & WER$\downarrow$ & MCD$\downarrow$ & $F_0$ Corr$\uparrow$ & PESQ$\uparrow$ & SI-SDR$\uparrow$ & STOI$\uparrow$ & $\uparrow$ \\
\midrule
OLIVE-A (Gain) [small 100k] & 0.0898 & 0.0726 & \underline{0.4840} & 0.6270 & 0.1280 & \textbf{7.3100} & \underline{0.2780} & \underline{2.8900} & \textbf{9.8100} & 0.9460 & 25.77 \\
OLIVE-A (Mix) [small 100k] & \underline{0.0863} & \textbf{0.0661} & \textbf{0.4950} & \textbf{0.6520} & \underline{0.1210} & 7.5400 & \underline{0.2780} & \textbf{3.0100} & 9.2600 & \textbf{0.9490} & \textbf{57.75} \\
OLIVE-A (Mix+Gain) [small 100k] & \textbf{0.0812} & \underline{0.0678} & 0.4730 & \underline{0.6380} & \textbf{0.1090} & \underline{7.5000} & \textbf{0.2820} & 2.8800 & \underline{9.3600} & \underline{0.9480} & \underline{52.01} \\
\bottomrule
\end{tabular}
}
\end{table*}

%% file: tables/hifigan_layer_ablation.tex
\begin{table*}[!ht]
\centering
\caption{Layer-wise HiFi-GAN conditioning ablation. Separate HiFi-GAN V2 vocoders were trained for 100k updates on frozen features from
OLIVE-A (Mix), OLIVE-A (Mix+Gain), and WavLM Base. ``Local encoder features'' denotes the output of the shared local encoder before the Transformer, and ``Transformer $k$'' denotes the output of the $k$th Transformer layer. \textbf{Bold}: best per model.}
\label{tab:hifigan_layer_ablation}
\resizebox{\textwidth}{!}{
\begin{tabular}{llcccccc}
\toprule
Model & Conditioning & Mel-L1$\downarrow$ & MCD$\downarrow$ & STOI$\uparrow$ & PESQ$\uparrow$ & $F_0$ MAE$\downarrow$ & UTMOS$\uparrow$ \\
\midrule
OLIVE-A (Mix) & Local encoder features & \textbf{0.5556} & \textbf{5.1394} & \textbf{0.9044} & \textbf{2.2269} & \textbf{13.2972} & \textbf{3.4484} \\
OLIVE-A (Mix) & Transformer 1 & 0.5951 & 5.5717 & 0.8759 & 1.8747 & 14.4624 & 3.1674 \\
OLIVE-A (Mix) & Transformer 2 & 0.6323 & 5.9658 & 0.8583 & 1.5139 & 14.9328 & 3.0849 \\
OLIVE-A (Mix) & Transformer 3 & 0.6461 & 5.9678 & 0.8415 & 1.4350 & 16.6105 & 3.0969 \\
OLIVE-A (Mix) & Transformer 4 & 0.6732 & 6.0818 & 0.8262 & 1.3332 & 19.4579 & 3.0786 \\
OLIVE-A (Mix) & Transformer 5 & 0.7036 & 6.3123 & 0.8070 & 1.2558 & 21.6921 & 2.9327 \\
\midrule
OLIVE-A (Mix+Gain) & Local encoder features & \textbf{0.5611} & \textbf{5.1712} & \textbf{0.9070} & \textbf{2.2426} & \textbf{12.6522} & \textbf{3.5567} \\
OLIVE-A (Mix+Gain) & Transformer 1 & 0.5947 & 5.5902 & 0.8802 & 1.8566 & 13.8737 & 3.3040 \\
OLIVE-A (Mix+Gain) & Transformer 2 & 0.6357 & 5.8769 & 0.8549 & 1.4910 & 16.3225 & 3.0375 \\
OLIVE-A (Mix+Gain) & Transformer 3 & 0.6316 & 5.8134 & 0.8618 & 1.5785 & 15.2443 & 3.2385 \\
OLIVE-A (Mix+Gain) & Transformer 4 & 0.6638 & 6.1259 & 0.8465 & 1.4499 & 16.6433 & 3.1341 \\
OLIVE-A (Mix+Gain) & Transformer 5 & 0.7110 & 6.4318 & 0.8121 & 1.3221 & 19.7228 & 2.9569 \\
\midrule
WavLM Base & Local encoder features & \textbf{0.5422} & \textbf{5.3733} & \textbf{0.8928} & \textbf{1.9573} & \textbf{13.9462} & \textbf{3.3709} \\
WavLM Base         & Transformer 1 & 0.5596 & 5.5364 & 0.8811 & 1.7441 & 14.9699 & 3.3503 \\
WavLM Base         & Transformer 2 & 0.6080 & 5.8671 & 0.8411 & 1.4173 & 17.3569 & 2.8881 \\
WavLM Base         & Transformer 3 & 0.6127 & 5.8017 & 0.8386 & 1.3218 & 20.3122 & 2.8329 \\
WavLM Base         & Transformer 4 & 0.6498 & 6.1132 & 0.8133 & 1.2944 & 24.1076 & 2.8239 \\
WavLM Base         & Transformer 5 & 0.6876 & 6.4009 & 0.7867 & 1.1658 & 26.2397 & 2.9338 \\
\bottomrule
\end{tabular}
}
\end{table*}

%% file: tables/hifigan_augmentation_ablation.tex
\begin{table}[!ht]
\centering
\caption{HiFi-GAN augmentation ablation for separate vocoder training. We
compare matched HiFi-GAN V2 vocoders trained on the same frozen
\textbf{OLIVE-A (Mix+Gain)} local encoder features, with and without waveform
mixup augmentation. \textbf{Bold}: better.}
\label{tab:hifigan_augmentation_ablation}
\begin{tabular}{lcccccc}
\toprule
Training data & Mel-L1$\downarrow$ & MCD$\downarrow$ & STOI$\uparrow$ & PESQ$\uparrow$ & $F_0$ MAE$\downarrow$ & UTMOS$\uparrow$ \\
\midrule
No augmentation & 0.8251 & 6.4177 & 0.7541 & 1.2384 & 16.9843 & 1.9400 \\
Waveform mixup & \textbf{0.5297} & \textbf{4.9983} & \textbf{0.9045} & \textbf{2.2566} & \textbf{13.0708} & \textbf{3.3836} \\
\bottomrule
\end{tabular}
\end{table}

%% file: figures/superb_three_radials.tex
\input{figures/superb_radar_macros}
\begin{tikzpicture}[font=\sffamily]
    \begin{scope}[shift={(-7.1,0)}]
        \TaskAxes{OLIVE variants vs.\ data2vec 2.0}
        \TaskRadarPlot{taskData2vec}{(90:{\TaskR*990/1000}) -- (64.286:{\TaskR*1000/1000}) -- (38.571:{\TaskR*994/1000}) -- (12.857:{\TaskR*666/1000}) -- (-12.857:{\TaskR*547/1000}) -- (-38.571:{\TaskR*1000/1000}) -- (-64.286:{\TaskR*714/1000}) -- (-90:{\TaskR*901/1000}) -- (-115.714:{\TaskR*999/1000}) -- (-141.429:{\TaskR*978/1000}) -- (-167.143:{\TaskR*1000/1000}) -- (-192.857:{\TaskR*574/1000}) -- (-218.571:{\TaskR*387/1000}) -- (-244.286:{\TaskR*926/1000})}
        \TaskRadarPlot{taskOliveMix}{(90:{\TaskR*967/1000}) -- (64.286:{\TaskR*913/1000}) -- (38.571:{\TaskR*997/1000}) -- (12.857:{\TaskR*451/1000}) -- (-12.857:{\TaskR*805/1000}) -- (-38.571:{\TaskR*904/1000}) -- (-64.286:{\TaskR*957/1000}) -- (-90:{\TaskR*982/1000}) -- (-115.714:{\TaskR*1000/1000}) -- (-141.429:{\TaskR*935/1000}) -- (-167.143:{\TaskR*914/1000}) -- (-192.857:{\TaskR*907/1000}) -- (-218.571:{\TaskR*968/1000}) -- (-244.286:{\TaskR*674/1000})}
        \TaskRadarPlot{taskOliveGain}{(90:{\TaskR*982/1000}) -- (64.286:{\TaskR*929/1000}) -- (38.571:{\TaskR*1000/1000}) -- (12.857:{\TaskR*540/1000}) -- (-12.857:{\TaskR*758/1000}) -- (-38.571:{\TaskR*1000/1000}) -- (-64.286:{\TaskR*1000/1000}) -- (-90:{\TaskR*1000/1000}) -- (-115.714:{\TaskR*998/1000}) -- (-141.429:{\TaskR*945/1000}) -- (-167.143:{\TaskR*934/1000}) -- (-192.857:{\TaskR*926/1000}) -- (-218.571:{\TaskR*968/1000}) -- (-244.286:{\TaskR*729/1000})}
        \TaskRadarPlot{taskOliveMask}{(90:{\TaskR*989/1000}) -- (64.286:{\TaskR*995/1000}) -- (38.571:{\TaskR*997/1000}) -- (12.857:{\TaskR*568/1000}) -- (-12.857:{\TaskR*777/1000}) -- (-38.571:{\TaskR*962/1000}) -- (-64.286:{\TaskR*643/1000}) -- (-90:{\TaskR*977/1000}) -- (-115.714:{\TaskR*995/1000}) -- (-141.429:{\TaskR*1000/1000}) -- (-167.143:{\TaskR*967/1000}) -- (-192.857:{\TaskR*481/1000}) -- (-218.571:{\TaskR*419/1000}) -- (-244.286:{\TaskR*906/1000})}
        \TaskRadarPlot{taskOliveJ}{(90:{\TaskR*968/1000}) -- (64.286:{\TaskR*913/1000}) -- (38.571:{\TaskR*1000/1000}) -- (12.857:{\TaskR*522/1000}) -- (-12.857:{\TaskR*1000/1000}) -- (-38.571:{\TaskR*942/1000}) -- (-64.286:{\TaskR*957/1000}) -- (-90:{\TaskR*919/1000}) -- (-115.714:{\TaskR*989/1000}) -- (-141.429:{\TaskR*926/1000}) -- (-167.143:{\TaskR*842/1000}) -- (-192.857:{\TaskR*944/1000}) -- (-218.571:{\TaskR*1000/1000}) -- (-244.286:{\TaskR*782/1000})}
    \end{scope}

    \begin{scope}[shift={(0,0)}]
        \TaskAxes{OLIVE-J variants vs.\ baselines}
        \TaskRadarPlot{taskWav2vec}{(90:{\TaskR*967/1000}) -- (64.286:{\TaskR*907/1000}) -- (38.571:{\TaskR*985/1000}) -- (12.857:{\TaskR*768/1000}) -- (-12.857:{\TaskR*800/1000}) -- (-38.571:{\TaskR*942/1000}) -- (-64.286:{\TaskR*671/1000}) -- (-90:{\TaskR*865/1000}) -- (-115.714:{\TaskR*931/1000}) -- (-141.429:{\TaskR*898/1000}) -- (-167.143:{\TaskR*809/1000}) -- (-192.857:{\TaskR*444/1000}) -- (-218.571:{\TaskR*355/1000}) -- (-244.286:{\TaskR*807/1000})}
        \TaskRadarPlot{taskHuBERT}{(90:{\TaskR*974/1000}) -- (64.286:{\TaskR*913/1000}) -- (38.571:{\TaskR*990/1000}) -- (12.857:{\TaskR*916/1000}) -- (-12.857:{\TaskR*810/1000}) -- (-38.571:{\TaskR*981/1000}) -- (-64.286:{\TaskR*729/1000}) -- (-90:{\TaskR*931/1000}) -- (-115.714:{\TaskR*990/1000}) -- (-141.429:{\TaskR*936/1000}) -- (-167.143:{\TaskR*868/1000}) -- (-192.857:{\TaskR*648/1000}) -- (-218.571:{\TaskR*226/1000}) -- (-244.286:{\TaskR*822/1000})}
        \TaskRadarPlot{taskWavLM}{(90:{\TaskR*977/1000}) -- (64.286:{\TaskR*923/1000}) -- (38.571:{\TaskR*997/1000}) -- (12.857:{\TaskR*1000/1000}) -- (-12.857:{\TaskR*749/1000}) -- (-38.571:{\TaskR*942/1000}) -- (-64.286:{\TaskR*843/1000}) -- (-90:{\TaskR*926/1000}) -- (-115.714:{\TaskR*994/1000}) -- (-141.429:{\TaskR*967/1000}) -- (-167.143:{\TaskR*901/1000}) -- (-192.857:{\TaskR*630/1000}) -- (-218.571:{\TaskR*516/1000}) -- (-244.286:{\TaskR*859/1000})}
        \TaskRadarPlot{taskData2vec}{(90:{\TaskR*990/1000}) -- (64.286:{\TaskR*1000/1000}) -- (38.571:{\TaskR*994/1000}) -- (12.857:{\TaskR*666/1000}) -- (-12.857:{\TaskR*547/1000}) -- (-38.571:{\TaskR*1000/1000}) -- (-64.286:{\TaskR*714/1000}) -- (-90:{\TaskR*901/1000}) -- (-115.714:{\TaskR*999/1000}) -- (-141.429:{\TaskR*978/1000}) -- (-167.143:{\TaskR*1000/1000}) -- (-192.857:{\TaskR*574/1000}) -- (-218.571:{\TaskR*387/1000}) -- (-244.286:{\TaskR*926/1000})}
        \TaskRadarPlot{taskOliveMask}{(90:{\TaskR*989/1000}) -- (64.286:{\TaskR*995/1000}) -- (38.571:{\TaskR*997/1000}) -- (12.857:{\TaskR*568/1000}) -- (-12.857:{\TaskR*777/1000}) -- (-38.571:{\TaskR*962/1000}) -- (-64.286:{\TaskR*643/1000}) -- (-90:{\TaskR*977/1000}) -- (-115.714:{\TaskR*995/1000}) -- (-141.429:{\TaskR*1000/1000}) -- (-167.143:{\TaskR*967/1000}) -- (-192.857:{\TaskR*481/1000}) -- (-218.571:{\TaskR*419/1000}) -- (-244.286:{\TaskR*906/1000})}
        \TaskRadarPlot{taskOliveJ}{(90:{\TaskR*968/1000}) -- (64.286:{\TaskR*913/1000}) -- (38.571:{\TaskR*1000/1000}) -- (12.857:{\TaskR*522/1000}) -- (-12.857:{\TaskR*1000/1000}) -- (-38.571:{\TaskR*942/1000}) -- (-64.286:{\TaskR*957/1000}) -- (-90:{\TaskR*919/1000}) -- (-115.714:{\TaskR*989/1000}) -- (-141.429:{\TaskR*926/1000}) -- (-167.143:{\TaskR*842/1000}) -- (-192.857:{\TaskR*944/1000}) -- (-218.571:{\TaskR*1000/1000}) -- (-244.286:{\TaskR*782/1000})}
    \end{scope}

    \begin{scope}[shift={(7.1,0)}]
        \CatAxes{OLIVE-J variants vs.\ baselines}
        \CatRadarPlot{taskWav2vec}{907}{805}{865}{880}{535}
        \CatRadarPlot{taskHuBERT}{948}{840}{931}{932}{565}
        \CatRadarPlot{taskWavLM}{974}{845}{926}{954}{668}
        \CatRadarPlot{taskData2vec}{913}{754}{901}{992}{629}
        \CatRadarPlot{taskOliveMask}{887}{794}{977}{987}{602}
        \CatRadarPlot{taskOliveJ}{851}{966}{919}{919}{909}
    \end{scope}

    \node[font=\small\sffamily\bfseries, align=center] at (0,-3.28) {
        \LegendItem{taskOliveJ}{OLIVE-J (Mix)}\quad
        \LegendItem{taskOliveMask}{OLIVE-J (Mask-only)}\quad
        \LegendItem{taskOliveMix}{OLIVE-A (Mix)}\quad
        \LegendItem{taskOliveGain}{OLIVE-A (Mix+Gain)}\quad
        \LegendItem{taskData2vec}{data2vec 2.0}\quad
        \LegendItem{taskWav2vec}{wav2vec 2.0}\quad
        \LegendItem{taskHuBERT}{HuBERT}\quad
        \LegendItem{taskWavLM}{WavLM}
    };
\end{tikzpicture}

%% file: tables/asr_lm_full.tex
\begin{table}[H]
\centering
\scriptsize
\caption{ASR fine-tuning results on LibriSpeech using \texttt{train-clean-100} as the labeled fine-tuning set. Word error rates (\%) are reported with and without a 4-gram language model; lower is better. \textbf{Bold}: best; \underline{underlined}: second-best.}
\label{tab:asr-lm-full}
\resizebox{\textwidth}{!}{%
\begin{tabular}{lrrrrrrrr}
\toprule
 & \multicolumn{4}{c}{No LM} & \multicolumn{4}{c}{4-gram LM} \\
\cmidrule(lr){2-5} \cmidrule(lr){6-9}
Model & dev-clean & dev-other & test-clean & test-other & dev-clean & dev-other & test-clean & test-other \\
\midrule
FBANK & 22.7 & 52.1 & 23.1 & 53.9 & 14.4 & 41.9 & 15.0 & 43.7 \\
wav2vec 2.0 Base & 6.0 & 16.3 & 6.5 & 16.0 & 4.1 & 12.6 & 4.9 & 12.6 \\
HuBERT Base & 5.9 & 15.5 & 6.4 & 15.6 & 4.1 & 12.1 & 4.7 & 12.0 \\
WavLM Base & 5.6 & 15.0 & 6.2 & 15.0 & 3.8 & 11.6 & 4.6 & 11.6 \\
data2vec Base & 4.7 & 12.0 & 5.0 & 12.3 & \underline{3.3} & 9.5 & \underline{3.9} & 9.7 \\
data2vec 2.0 Base & \textbf{4.5} & \textbf{11.0} & \textbf{4.8} & \textbf{11.1} & \textbf{3.2} & \textbf{8.6} & \textbf{3.8} & \textbf{8.8} \\
\midrule
\rowcolor{olivehighlight}
OLIVE-A (Mix) & 5.9 & 14.7 & 6.4 & 14.4 & 4.2 & 11.4 & 4.7 & 11.2 \\
\rowcolor{olivehighlight}
OLIVE-A (Mix+Gain) & 5.6 & 14.0 & 6.1 & 14.0 & 3.8 & 11.0 & 4.6 & 11.0 \\
\rowcolor{olivehighlight}
OLIVE-J (Mask-only) & \textbf{4.5} & \underline{11.4} & \underline{4.9} & \underline{11.4} & \underline{3.3} & \underline{9.0} & \underline{3.9} & \underline{9.0} \\
\rowcolor{olivehighlight}
OLIVE-J (Mix) & 6.2 & 15.6 & 6.4 & 15.7 & 4.3 & 12.1 & 5.0 & 12.2 \\
\bottomrule
\end{tabular}
}
\end{table}

%% file: tables/resynthesis_wer_secs.tex
\begin{table}[H]
\centering
\setlength{\tabcolsep}{6pt}
\scriptsize
\caption{Intelligibility and speaker preservation of resynthesized speech on all 2{,}620 LibriSpeech test-clean utterances, using the same vocoders as Table~\ref{tab:resynthesis_main}. WER is computed with Whisper large-v3; SECS is the mean cosine similarity between ECAPA-TDNN speaker embeddings of the resynthesized and reference utterances. \textbf{Bold}: best; \underline{underlined}: second-best.}
\label{tab:resynthesis_wer_secs}
\begin{tabular}{lcc}
\toprule
Model & WER (\%) $\downarrow$ & SECS $\uparrow$ \\
\midrule
Mel spectrogram & 4.23 & 0.565 \\
wav2vec 2.0 Base & 3.10 & 0.645 \\
HuBERT Base & 2.96 & 0.668 \\
WavLM Base & 3.06 & 0.670 \\
data2vec Base & 3.09 & 0.664 \\
data2vec 2.0 Base & 3.01 & 0.676 \\
\midrule
\rowcolor{olivehighlight}
\multicolumn{3}{l}{\emph{Frozen feature-conditioned vocoder}} \\
\rowcolor{olivehighlight}
OLIVE-A (Mix+Gain) & 3.05 & 0.696 \\
\rowcolor{olivehighlight}
OLIVE-J (Mask-only) & \underline{2.95} & 0.709 \\
\rowcolor{olivehighlight}
OLIVE-J (Mix) & \textbf{2.92} & \textbf{0.736} \\
\midrule
\rowcolor{oliveintegrated}
\multicolumn{3}{l}{\emph{Integrated vocoder}} \\
\rowcolor{oliveintegrated}
OLIVE-J (Mask-only) & 3.07 & 0.693 \\
\rowcolor{oliveintegrated}
OLIVE-J (Mix) & 3.04 & \underline{0.722} \\
\bottomrule
\end{tabular}
\end{table}

%% file: tables/hifigan.tex
\begin{table}[H]
\centering
\setlength{\tabcolsep}{4pt}
\scriptsize
\caption{Waveform reconstruction on LibriSpeech test-clean with matched and integrated HiFi-GAN V2 vocoders: spectral and pitch metrics. Values are mean $\pm$ margin of error from 95\% bootstrap confidence intervals. \textbf{Bold}: best; \underline{underlined}: second-best.}
\label{tab:resynthesis_test_clean}
\begin{tabular}{lcccc}
\toprule
Model & Mel-L1 $\downarrow$ & MCD $\downarrow$ & LSD $\downarrow$ & $F_0$ MAE (Hz) $\downarrow$ \\
\midrule
Mel spectrogram & 0.677 $\pm$ 0.004 & 6.24 $\pm$ 0.03 & 0.829 $\pm$ 0.005 & 14.4 $\pm$ 0.3 \\
wav2vec 2.0 Base & 0.474 $\pm$ 0.003 & 5.18 $\pm$ 0.03 & 0.604 $\pm$ 0.004 & 11.6 $\pm$ 0.3 \\
HuBERT Base & 0.475 $\pm$ 0.002 & 5.05 $\pm$ 0.02 & 0.604 $\pm$ 0.003 & 13.3 $\pm$ 0.3 \\
WavLM Base & 0.470 $\pm$ 0.002 & 5.01 $\pm$ 0.02 & 0.599 $\pm$ 0.003 & 13.1 $\pm$ 0.3 \\
data2vec Base & 0.467 $\pm$ 0.002 & 4.90 $\pm$ 0.02 & 0.594 $\pm$ 0.003 & 12.8 $\pm$ 0.3 \\
data2vec 2.0 Base & 0.434 $\pm$ 0.002 & 4.78 $\pm$ 0.02 & 0.558 $\pm$ 0.003 & 13.0 $\pm$ 0.3 \\
\midrule
\rowcolor{olivehighlight}
\multicolumn{5}{l}{\emph{Frozen feature-conditioned vocoder}} \\
\rowcolor{olivehighlight}
OLIVE-A (Mix+Gain) & 0.446 $\pm$ 0.002 & 4.69 $\pm$ 0.02 & 0.566 $\pm$ 0.003 & 11.6 $\pm$ 0.2 \\
\rowcolor{olivehighlight}
OLIVE-J (Mask-only) & \textbf{0.410 $\pm$ 0.002} & \underline{4.43 $\pm$ 0.01} & \textbf{0.525 $\pm$ 0.002} & 10.8 $\pm$ 0.2 \\
\rowcolor{olivehighlight}
OLIVE-J (Mix) & \underline{0.421 $\pm$ 0.002} & \textbf{4.35 $\pm$ 0.02} & \underline{0.528 $\pm$ 0.002} & \textbf{9.6 $\pm$ 0.2} \\
\midrule
\rowcolor{oliveintegrated}
\multicolumn{5}{l}{\emph{Integrated vocoder}} \\
\rowcolor{oliveintegrated}
OLIVE-J (Mask-only) & 0.476 $\pm$ 0.002 & 4.68 $\pm$ 0.01 & 0.593 $\pm$ 0.002 & 11.1 $\pm$ 0.3 \\
\rowcolor{oliveintegrated}
OLIVE-J (Mix) & 0.579 $\pm$ 0.003 & 4.56 $\pm$ 0.02 & 0.688 $\pm$ 0.003 & \underline{10.0 $\pm$ 0.2} \\
\bottomrule
\end{tabular}
\end{table}

\begin{table}[H]
\centering
\setlength{\tabcolsep}{4pt}
\scriptsize
\caption{Waveform reconstruction on LibriSpeech test-clean with matched and integrated HiFi-GAN V2 vocoders: intelligibility, perceptual, and signal-level metrics. Values are mean $\pm$ margin of error from 95\% bootstrap confidence intervals. \textbf{Bold}: best; \underline{underlined}: second-best.}
\label{tab:resynthesis_test_clean_waveform}
\begin{tabular}{lcccccc}
\toprule
Model & STOI $\uparrow$ & PESQ $\uparrow$ & ViSQOL $\uparrow$ & SI-SDR $\uparrow$ & SNR $\uparrow$ & UTMOS $\uparrow$ \\
\midrule
Reference audio &  &  &  &  &  & 4.09 $\pm$ 0.01 \\
\midrule
Mel spectrogram & 0.833 $\pm$ 0.001 & 1.42 $\pm$ 0.01 & 3.59 $\pm$ 0.01 & -18.02 $\pm$ 0.15 & -2.60 $\pm$ 0.02 & 2.98 $\pm$ 0.02 \\
wav2vec 2.0 Base & 0.912 $\pm$ 0.001 & 2.19 $\pm$ 0.01 & 4.04 $\pm$ 0.01 & -7.53 $\pm$ 0.21 & -0.42 $\pm$ 0.06 & 3.58 $\pm$ 0.02 \\
HuBERT Base & 0.914 $\pm$ 0.001 & 2.11 $\pm$ 0.01 & 4.05 $\pm$ 0.01 & -11.78 $\pm$ 0.17 & -1.57 $\pm$ 0.04 & 3.63 $\pm$ 0.02 \\
WavLM Base & 0.917 $\pm$ 0.001 & 2.21 $\pm$ 0.01 & 4.05 $\pm$ 0.01 & -10.93 $\pm$ 0.20 & -1.31 $\pm$ 0.04 & 3.68 $\pm$ 0.02 \\
data2vec Base & 0.902 $\pm$ 0.001 & 2.10 $\pm$ 0.01 & 4.06 $\pm$ 0.01 & -10.64 $\pm$ 0.21 & -1.15 $\pm$ 0.04 & 3.50 $\pm$ 0.02 \\
data2vec 2.0 Base & 0.917 $\pm$ 0.001 & 2.21 $\pm$ 0.01 & 4.10 $\pm$ 0.01 & -9.67 $\pm$ 0.18 & -1.11 $\pm$ 0.04 & 3.63 $\pm$ 0.02 \\
\midrule
\rowcolor{olivehighlight}
\multicolumn{7}{l}{\emph{Frozen feature-conditioned vocoder}} \\
\rowcolor{olivehighlight}
OLIVE-A (Mix+Gain) & 0.929 $\pm$ 0.001 & 2.62 $\pm$ 0.01 & 4.18 $\pm$ 0.01 & -3.36 $\pm$ 0.18 & 1.06 $\pm$ 0.08 & 3.70 $\pm$ 0.02 \\
\rowcolor{olivehighlight}
OLIVE-J (Mask-only) & \underline{0.935 $\pm$ 0.001} & 2.81 $\pm$ 0.01 & \underline{4.26 $\pm$ 0.01} & 0.97 $\pm$ 0.15 & 3.31 $\pm$ 0.08 & \textbf{3.86 $\pm$ 0.01} \\
\rowcolor{olivehighlight}
OLIVE-J (Mix) & \textbf{0.942 $\pm$ 0.001} & \textbf{3.06 $\pm$ 0.01} & \textbf{4.34 $\pm$ 0.01} & \textbf{2.94 $\pm$ 0.16} & \textbf{4.69 $\pm$ 0.10} & \underline{3.83 $\pm$ 0.02} \\
\midrule
\rowcolor{oliveintegrated}
\multicolumn{7}{l}{\emph{Integrated vocoder}} \\
\rowcolor{oliveintegrated}
OLIVE-J (Mask-only) & 0.911 $\pm$ 0.001 & 2.56 $\pm$ 0.01 & 4.14 $\pm$ 0.01 & 0.72 $\pm$ 0.14 & 3.08 $\pm$ 0.08 & 3.74 $\pm$ 0.02 \\
\rowcolor{oliveintegrated}
OLIVE-J (Mix) & 0.921 $\pm$ 0.001 & \underline{2.88 $\pm$ 0.02} & 4.21 $\pm$ 0.01 & \underline{2.50 $\pm$ 0.14} & \underline{4.25 $\pm$ 0.09} & 3.78 $\pm$ 0.02 \\
\bottomrule
\end{tabular}
\end{table}